\documentclass[lettersize,journal]{IEEEtran}

\usepackage{CJKutf8}
\usepackage{amsmath}
\usepackage{amssymb}
\usepackage{algorithm}
\usepackage{enumitem}
\usepackage{picinpar}
\usepackage[noend]{algpseudocode}
\usepackage{graphicx}
\usepackage{caption}

\usepackage{algorithmicx}
\usepackage{subfigure}
\usepackage{multirow}
\usepackage{color}
\usepackage{xcolor}
\usepackage{balance}
\usepackage{enumitem}
\usepackage{hhline}
\usepackage[normalem]{ulem}
\usepackage{booktabs}
\usepackage{wrapfig}
\usepackage{cancel}
\usepackage{hyperref}
\usepackage{makecell}
\usepackage[switch,columnwise]{lineno}

\newcommand{\bL}{\ensuremath{\mathcal{L}}}

\newcommand{\bS}{\ensuremath{\mathcal{S}}}

\newcommand{\bV}{\ensuremath{\mathcal{V}}}

\newcommand{\bN}{\ensuremath{\mathcal{N}}}

\newcommand{\bT}{\ensuremath{\mathcal{T}}}
\newcommand{\bQ}{\ensuremath{\mathcal{Q}}}

\renewcommand{\vec}[1]{\ensuremath{\mathbf{#1}}}

\newcommand{\stitle}[1]{\vspace{1mm} \noindent {\bf #1}}

\newcommand{\eg}{{\it e.g.}}
\newcommand{\etal}{{\it et al.}}
\newcommand{\ie}{{\it i.e.}}

\newcommand{\wrt}{w.r.t. }

\newcommand{\eat}[1]{}

\newcommand{\stkout}[1]{\ifmmode\text{\sout{\ensuremath{#1}}}\else\sout{#1}\fi}

\AtBeginDocument{%
  }

\begin{document}

\title{A Survey of Few-Shot Learning on Graphs: from Meta-Learning to Pre-Training and Prompt Learning}

\author{Xingtong~Yu,
        Yuan~Fang,~\IEEEmembership{Senior Member,~IEEE,}
        Zemin~Liu,
        Yuxia~Wu,
        Zhihao~Wen,
        Jianyuan~Bo,
        Xinming~Zhang,~\IEEEmembership{Senior Member,~IEEE,}
        and Steven C.H. Hoi,~\IEEEmembership{Fellow,~IEEE,}
        
\IEEEcompsocitemizethanks{
\IEEEcompsocthanksitem Xingtong Yu, Yuan Fang, Yuxia Wu, Zhihao Wen, Jianyuan Bo, Xinming Zhang and Steven C.H. Hoi are with Singapore Management University, Singapore 178902 (emails: \{xingtongyu, yfang, yuxiawu, zhwen.2019, jybo.2020, chhoi\}@smu.edu.sg). 
\IEEEcompsocthanksitem Zemin Liu is with the Zhejiang University, Hangzhou, Zhejiang 310058 (e-mail: liu.zemin@zju.edu.cn).
\IEEEcompsocthanksitem Xinming Zhang is with the University of Science and Technology of China, Hefei, Anhui 230052, China (email: xinming@ustc.edu.cn).
}
\thanks{Co-first author: Xingtong Yu; Yuan Fang. Corresponding author: Yuan Fang.}
\thanks{Manuscript received April 19, 2021; revised August 16, 2021.}}

\markboth{Journal of \LaTeX\ Class Files,~Vol.~14, No.~8, August~2021}%
{Yu \MakeLowercase{\textit{et al.}}: A Survey of Few-Shot Learning on Graphs: Meta-Learning, Pre-Training, Prompting and Beyond}


\IEEEtitleabstractindextext{%
\begin{abstract}
Graph representation learning, a critical step in graph-centric tasks, has seen significant advancements. Earlier techniques often operate in an end-to-end setting, which heavily rely on the availability of ample labeled data. This constraint has spurred the emergence of \emph{few-shot learning on graphs}, where only a few labels are available for each task. Given the extensive literature in this field, this survey endeavors to synthesize recent developments, provide comparative insights, and identify future directions. We systematically categorize existing studies based on two major taxonomies: (1) Problem taxonomy, which explores different types of data scarcity problems and their applications, and (2) Technique taxonomy, which details key strategies for addressing these data-scarce few-shot problems. The techniques can be broadly categorized into meta-learning, pre-training, and hybrid approaches, with a finer-grained classification in each category to aid readers in their method selection process. Within each category, we analyze the relationships among these methods and compare their strengths and limitations. Finally, we outline prospective directions for few-shot learning on graphs to catalyze continued innovation in this field. The website for this survey can be accessed by \url{https://github.com/smufang/fewshotgraph}.
\end{abstract}

}
\maketitle

\IEEEdisplaynontitleabstractindextext

%
\IEEEpeerreviewmaketitle

\section{Introduction}\label{sec.intro}
Data objects are often involved in complex interactions with each other, forming diverse network structures also known as \emph{graphs}. The prevalence of graph-structured data has led to a growing demand in graph-centric tasks, such as node classification \cite{kipf2016semi}, link prediction \cite{zhang2018link}, and graph classification \cite{zhang2018end}. 
Conventional graph analysis usually relies on feature engineering to exploit the structural information  \cite{page1999pagerank,jeh2002simrank,backstrom2011supervised}, necessitating significant manual effort. The emergence of \emph{graph embedding}  \cite{perozzi2014deepwalk,grover2016node2vec,tang2015line} opens up great opportunities for automated \emph{graph representation learning} without requiring handcrafted features. Generally, these approaches encode various graph elements (\eg, nodes, edges or subgraphs) into low-dimensional vectors. By leveraging co-occurrence patterns among these elements, they aim to preserve graph structural information in a latent vector space \cite{cai2018comprehensive}.
Subsequently, graph neural networks (GNNs) \cite{kipf2016semi,velivckovic2017graph,yu2023learning} have
integrated both structural information and node semantic features, achieving state-of-the-art performance in graph-centric tasks.
Modern GNNs typically adopt a message-passing framework, where each node receives and aggregate features from neighboring nodes recursively. As a result, both structural and semantic information are incorporated  into graph representations, empowering GNNs to attain state-of-the-art performance in various downstream tasks \cite{wu2020comprehensive}.
More recently, given the remarkable success of transformers in the fields of natural language processing (NLP) and computer vision (CV), transformer-based architectures \cite{yun2019graph,hu2020heterogeneous,rampavsek2022recipe,kim2022pure,ying2021transformers} have also gained popularity in graph learning.

\begin{figure}[t]
\centering
\includegraphics[width=1\linewidth]{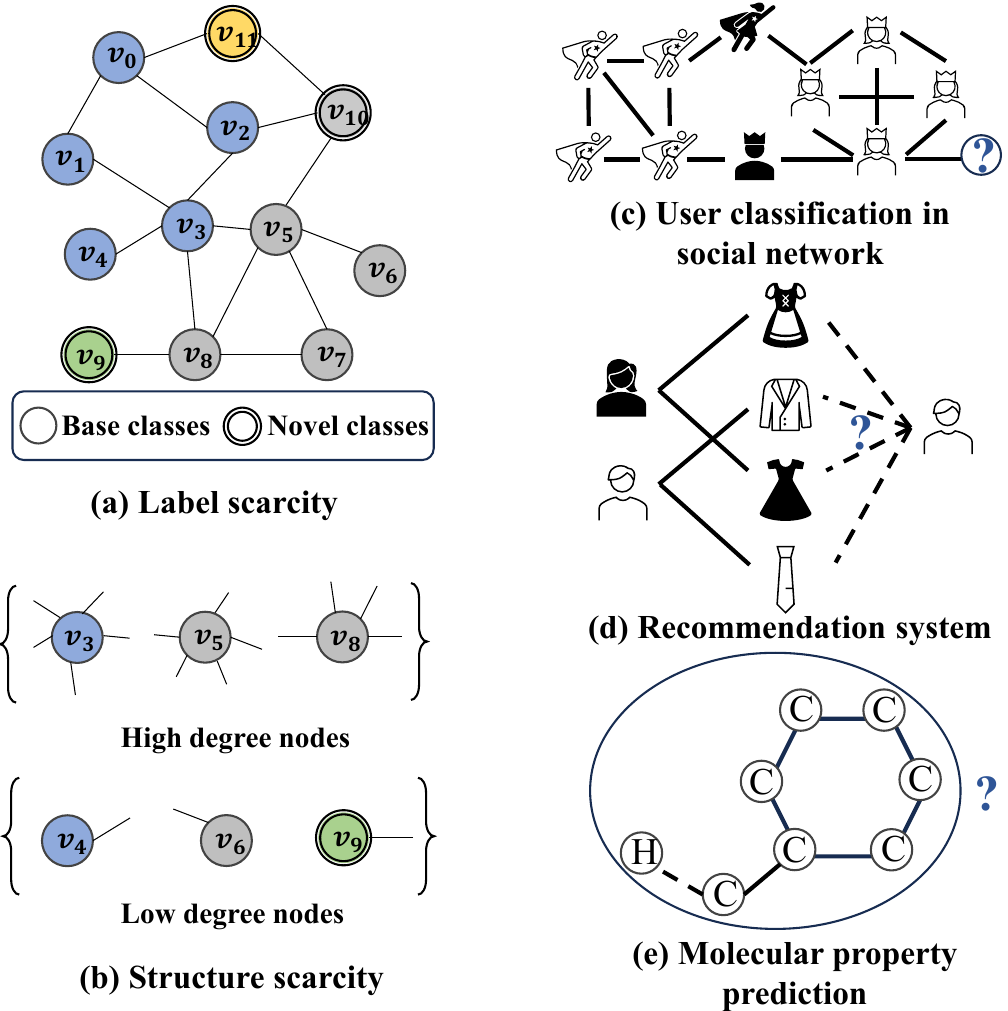}
\caption{Overview of few-shot learning on graphs. (a--b) Different problem settings involving label or structure scarcity on graphs. (c--e) Different applications, such as cold-start recommendation on user-item graphs.}
\label{fig.problem-application}
\end{figure}

\begin{figure}[t]
\centering
\includegraphics[width=0.8\linewidth]{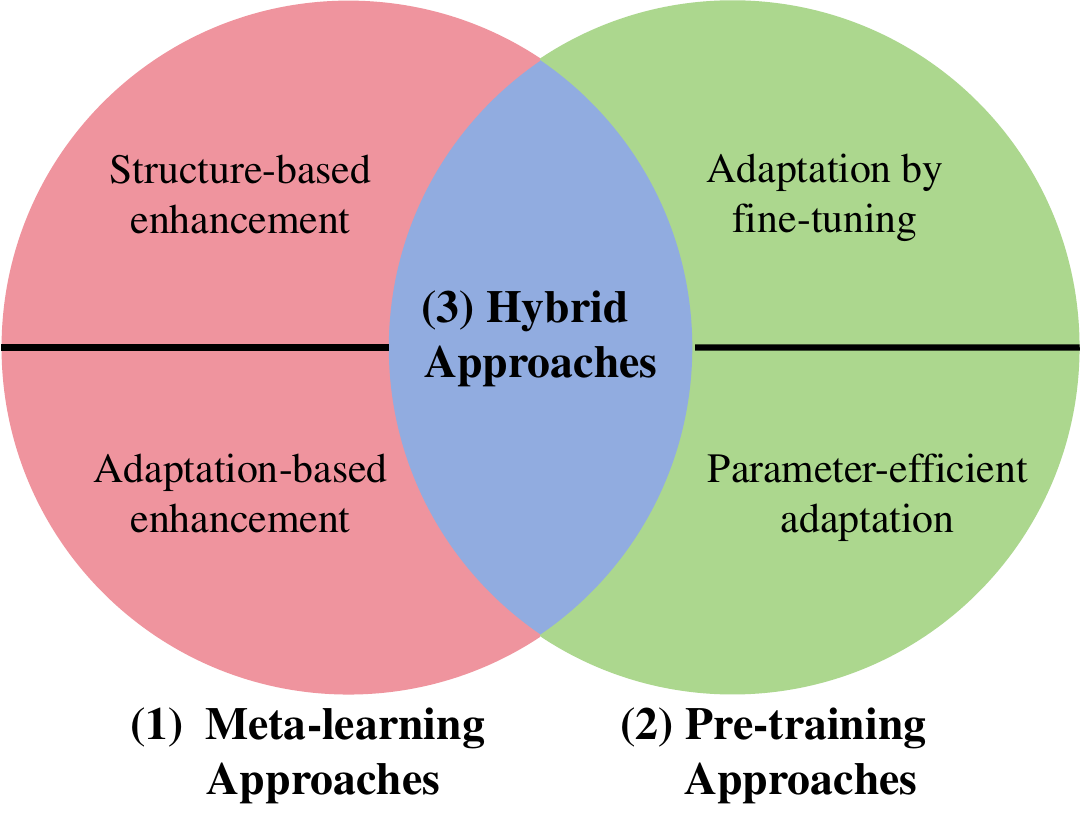} 
\vspace{-2mm}
\caption{Overview of few-shot learning techniques on graphs.}
\label{fig.intro-framework}
\end{figure}

\subsection{Few-Shot Learning on Graphs}
Unfortunately, the effectiveness of prevailing graph representation learning techniques, including GNNs and graph transformers, depends on not only rich graph structures, but also extensive labeled data. However, data scarcity is common in real-world applications, significantly hindering their performance. Specifically, we observe data scarcity in two broad categories: (1) \emph{Label} scarcity and (2) \emph{Structure} scarcity. First, acquiring labeled data is often challenging or expensive \cite{liu2023graphprompt}, leading to label scarcity, especially for novel classes, as illustrated in Fig.~\ref{fig.problem-application}(a). Second, graph structures may be sparse, especially for a large number of low-degree  nodes with a limited number of neighbors, as shown in Fig.~\ref{fig.problem-application}(b), resulting in structure scarcity.

In light of data scarcity, \emph{few-shot learning} methods on graphs \cite{garcia2018fewshot,guo2021few}  have attracted significant attention across various domains, such as social network analysis \cite{ding2021few,qian2021adapting}, recommendation systems \cite{li2020few,lu2020meta}, and molecular research \cite{guo2021few,wang2021property}, as shown in Fig.~\ref{fig.problem-application}(c,d,e). These approaches boil down to learning universal, task-independent prior knowledge from graphs, which is then adapted to tackle tasks with label scarcity or graphs with structure scarcity. Based on the specific strategies for learning prior knowledge and downstream adaptation, we further classify these techniques into three major families: (1) \emph{Meta-learning} approaches, (2) \emph{Pre-training} approaches, and (3) \emph{Hybrid} approaches, as depicted in Fig.~\ref{fig.intro-framework}. Notably, both meta-learning \cite{ding2020graph,sheng2020adaptive} and pre-training \cite{velivckovic2018deep,qiu2020gcc} represent two predominant research trends for learning priors from additional data, which are then adapted for few-shot downstream tasks. However, these two trends follow distinct paradigms, as we further elaborate next.


\stitle{Meta-learning on graphs.}
Mainstream meta-learning approaches, such as Model Agnostic Meta-Learning (MAML) \cite{finn2017model} and Prototypical Networks (Protonets) \cite{snell2017prototypical}, derive prior knowledge from a series of ``meta-training'' tasks (with task-specific labels) that mirror the downstream ``meta-testing'' tasks, assuming that all these tasks are independently drawn from an identical task distribution. 
Instead of directly learning a model, they aim to learn model initialization parameters or adaptation mechanisms that can be rapidly fine-tuned for new tasks---a concept known as ``learning to learn.''

Meta-learning presents a compelling approach to few-shot learning on graphs. 
We introduce a novel taxonomy that focuses on the diverse ways in which graph data enhance existing meta-learning frameworks. 
Specifically, we identify two major forms of graph-enabled enhancement: \textit{structure}-based and \textit{adaptation}-based enhancements. On one hand, structure-based enhancement utilizes additional contexts provided by graphs \cite{yin2017local,xiao2021hmnet}, categorized further into node-, edge-, and subgraph/graph-level contexts. These graph-grounded contexts provide complementary signals at various granularity levels, supplementing existing meta-learning methods for few-shot learning. On the other hand, adaptation-based enhancement focuses on leveraging graphs to enhance or refine the adaptation step \cite{suo2020tadanet,yao2020graph}. The adaptation step, crucial in many meta-learning frameworks, involves adapting the learned meta-knowledge to new tasks during testing. In particular, adaptation can be enhanced in various ways, including at the task, graph, and feature levels.

\stitle{Pre-training on graphs.}
On the other hand, pre-training approaches utilize unlabeled data to optimize self-supervised pretext tasks, generally unrelated to the downstream tasks. They directly train a model as the prior, followed by an additional tuning phase to adapt the prior to specific downstream tasks \cite{kenton2019bert,chen2020simple}.  
As pre-training emerges as a pivotal strategy for few-shot learning on graphs, we categorize these methods based on the specific techniques utilized during the pre-training and downstream adaptation phases.

In the pre-training phase, pre-training approaches commonly train a graph encoder (i.e., the prior), aiming to capture inherent characteristics of graph data, such as local connectivity \cite{hu2020gpt,qiu2020gcc} or global topological patterns \cite{you2020graph,you2021graph}. These approaches adopt self-supervised pretext tasks that are either \textit{contrastive} or \textit{generative}. Specifically,  contrastive strategies \cite{velivckovic2018deep,sun2019infograph} derive meaningful representations by contrasting positive and negative pairs of instances, 
whereas generative strategies \cite{kipf2016variational,hu2020gpt} train graph encoders and decoders to reconstruct graph elements, such as masked node features or dropped edges. 
Apart from pre-training methods that focus solely on graph data, recent developments \cite{wen2023augmenting,liu2023one} have integrated language models into graph pre-training to exploit textual descriptions associated with graphs. 

Following the pre-training phase, in the downstream adaptation phase, pre-training approaches either perform a \emph{fine-tuning} of the pre-trained graph encoder on each task, or employ a more \emph{parameter-efficient} adaptation strategy. 
Generally, fine-tuning \cite{velivckovic2018deep,qiu2020gcc} updates the parameters in the pre-trained graph encoder and a task head based on labeled samples for each downstream task, thereby tailoring the pre-trained knowledge to each specific task. However, full fine-tuning can be expensive and often leads to overfitting in few-shot settings, where each task has only a few labeled samples.
Alternatively, \emph{parameter-efficient} adaptation freezes most or all of the parameters of the pre-trained model and tunes only a much smaller module.   
Specifically, parameter-efficient fine-tuning \cite{li2023adaptergnn,gui2023g} only updates a minimal subset of the pre-trained weights and the task head, or an additional lightweight module injected into the pre-trained model. 
In contrast, prompt-based learning \cite{sun2022gppt,liu2023graphprompt,sun2023all} 
adopts a unified template that bridges pretext and downstream tasks, eliminating the need for any task head. Subsequently, it learns a task-specific prompt, consisting of only a small number of parameters, for each downstream task while freezing the pre-trained graph encoder.

\stitle{Hybrid approaches.}
Finally, hybrid approaches that integrate both paradigms have emerged recently, thereby synergistically harnessing the strengths of meta-learning and pre-training approaches. These methods are able to exploit both unlabeled graph data and annotations from similar tasks, enabling more seamless adaptation to various downstream tasks.

\subsection{Differences with Existing Surveys and Our Contributions}
Over the past few years, many efforts have been devoted to reviewing the few-shot learning literature from diverse perspectives \cite{wang2020generalizing,song2023comprehensive,wang2019few}, but they do not cover few-shot learning on graphs.
Recently, Zhang \etal~\cite{zhang2022few} have discussed this field but limited its categorization to node-, edge- and graph-level problems in the meta-learning paradigm, lacking a comprehensive review of the data scarcity problems and technical approaches. Besides, there are also some related reviews that have surveyed meta-learning on graphs \cite{mandal2021meta}, pre-training on graphs \cite{xia2022survey} and prompting on graphs \cite{wu2023survey,sun2023graph}.  However, these surveys have predominantly focused on specific techniques for a single paradigm, tailored solely to the label scarcity problem, resulting in a relatively narrow coverage of the field.

Our survey endeavors to fill this gap by presenting a holistic overview of few-shot learning on graphs, proposing a \emph{multi-perspective and fine-grained} taxonomy of both problems and methodologies. 
More specifically, we systematically categorize the literature into two novel taxonomies based on their problem settings and technical approaches, respectively. We divide the problem taxonomy into \emph{label} scarcity and \emph{structure} scarcity settings, while organizing the technique taxonomy into \emph{meta-learning}, \emph{pre-training}, and \emph{hybrid} approaches. Each branch is further examined and subdivided into finer categories.


In summary, the contributions of our survey are threefold, as follows.
    (1) We introduce two taxonomies for few-shot learning on graphs in terms of problem settings and technical methodologies, respectively, providing a systematic and comprehensive overview of the literature in the field. 
    (2) We synthesize recent progress within each branch of the taxonomies, offering comparative analyses of their respective strengths and weaknesses.
    (3) We discuss open challenges and outline promising directions for future research.

The remainder of this survey is organized as follows. Section~\ref{sec.background} introduces the relevant preliminaries and definitions used throughout the survey. Section~\ref{sec.pro} presents the taxonomy on data scarcity problems, discussing the issues of label scarcity and structure scarcity in the context of few-shot learning on graphs. Sections~\ref{sec.meta-learning}--\ref{sec.hybrid} delve into the finer categories of the taxonomy on methodologies. More specifically, Sections~\ref{sec.meta-learning} and~\ref{sec.pre-training} cover the meta-learning and pre-training paradigms, respectively, while Section~\ref{sec.hybrid} explores the hybrid approaches that combine both meta-learning and pre-training paradigms to leverage their respective advantages. Section~\ref{sec.future} outlines open challenges in the field and suggests potential directions for future research. Finally, Section~\ref{sec.conclusion} concludes the survey.

\section{Background}\label{sec.background}
In this section, we introduce the preliminaries for graph data and graph representation learning. 
We summarize the frequently used notations in Table~\ref{table.notation}.

\begin{table}[t]
\centering
\small
\addtolength{\tabcolsep}{-0pt}
\caption{Notations used in this paper.} 
\begin{tabular}{c | l }
\toprule
 \textbf{Notation} & \textbf{Description} \\
 \midrule
 $G$  & A graph $G$. \\ 
 $V$, $E$ & Node set $\bV$, edge set $E$. \\ 
 $\vec{X}_V$, $\vec{X}_E$ & The feature matrices of nodes and edges. \\ 
 $T$, $R$ & The set of node types and edge types. \\ 
 $\phi$, $\varphi$ &  Type mapping functions for nodes/edges. \\ 
 $\vec{x}_v$ & The attribute of node $v$.\\
 $\vec{h}_v$ ($\vec{h}^l_v$) & The representation of node $v$ (in layer $l$). \\  
 $\bN_v$ & Neighbor set of node $v$. \\  
 $\theta$ & The parameters set of a model. \\ 
 $\omega$ & The meta-knowledge. \\ 
 $\bT^i_{*}=(\bS^i_{*},\bQ^i_{*})$ & Task $i$, support set $\bS^i_{*}$, query set $\bQ^i_{*}$. \\
 $\mathcal{P}_o$ & Positive samples of target instance $o$.\\
 $\mathcal{N}_o$ & Negative samples of target instance $o$.\\ 
 $S_v$ & Subgraph of ego-node $v$. \\
 $\Vec{s}_v$ & Embedding of $S_v$. \\
 \bottomrule
\end{tabular}
\label{table.notation}
\end{table}

\subsection{Graph Formalization}

A \textbf{graph} is defined as $G=\{V,E,\vec{X}_V, \vec{X}_E, T,R,\phi,\varphi\}$, where $V$ is the set of nodes, $E$ is the set of edges, and $\vec{X}_V\in\mathbb{R}^{|V|\times d_V},\vec{X}_E\in\mathbb{R}^{|E|\times d_E}$ are the feature matrices of nodes and edges, respectively. Furthermore, $T$ and $R$ represent the set of node types and edge types, respectively. $\phi:V\rightarrow T$ is a mapping function that assigns each node $v\in V$ to its respective node type $\phi(v)\in T$, while  $\varphi:E\rightarrow R$ is another mapping function that assigns each edge $e\in E$ to its corresponding edge type $\varphi(e)\in R$.

In different applications, graphs often manifest in diverse forms to preserve application-specific connectivity information. 
In its most basic form, a \textbf{homogeneous graph} $G_{\text{homo}}$ is a graph where $|T|=1$ and $|R|=1$, meaning that it contains only one type of node and edge. 
Conversely, a \textbf{heterogeneous graph} $G_{\text{hete}}$ \cite{wang2019heterogeneous}, is a graph that satisfies $|T|>1$ or $|R|>1$. 
The rich semantic information conveyed by multiple node or edge types enables heterogeneous graphs to model many real-world scenarios \cite{dong2017metapath2vec,fu2017hin2vec}.  In particular, knowledge graphs\cite{fensel2020introduction} can be regarded as a form of heterogeneous graphs.
Moreover, graphs often evolve over time in real-world applications. A \textbf{dynamic graph} 
${G}_\text{dyn}=\{G^{(t)}:t\in \mathcal{T}\}$ 
models such scenarios, where $G^{(t)}$ represents the graph at time $t\in \mathcal{T}$ for some time domain $\mathcal{T}$, where the nodes, edges or their features can change over time.
Additionally, the nodes or edges can be associated with multi-modal content, such as numerical, textual and visual features \cite{wen2023augmenting,cao2022otkge}, forming a \textbf{multi-modal graph} $G_{\text{mm}}$ where $\vec{X}_V$ or $\vec{X}_E$ is a feature tensor for the nodes or edges, respectively. In particular, if each node or edge is associated with textual content, we also call it a \textbf{text-attributed graph}.


\subsection{Graph Representation Learning}
Graph representation learning \cite{cai2018comprehensive} has emerged as a mainstream technique for graph analysis. It embeds graph structures into a low-dimensional space while preserving their inherent connectivity patterns.
Formally, define $\vec{h}_v\in\mathbb{R}^d$ as the representation of node $v\in V$, formulated as:
\begin{equation}
    \vec{h}_v = f_g(v,G;\theta_g),
\end{equation}
where $f_g(\cdot;\theta_g)$ parameterized by $\theta_g$ is the representation learning function, also known as the \emph{graph encoder}, and $d$ is the dimension of the embedding vector $\vec{h}_v$.

Early graph representation learning approaches \cite{perozzi2014deepwalk,grover2016node2vec,tang2015line} usually exploit the co-occurrence relationships or proximity between nodes or substructures on graphs.
More recently, graph neural networks (GNNs) capitalize on a message-passing framework, in which each node derives its representation by receiving and aggregating messages (\ie, features) from neighboring nodes, thus capturing both structural and semantic information. Multiple message-passing layers can be stacked such that, in the $l$-th layer, we obtain the representation $\vec{h}^l_v\in\mathbb{R}^{d_l}$ of node $v$ as
\begin{align}
    \vec{h}^l_v = \textsc{Aggr}(\vec{h}^{l-1}_v,\{\vec{h}^{l-1}_u:u\in\bN_v\};\theta^l_g),
\end{align}
where $\bN_v$ is the set of neighbors of $v$, $\theta^l_g$ denotes the learnable parameters in the $l$-th layer, and $\textsc{Aggr}(\cdot)$ is a neighborhood aggregation function \cite{hamilton2017inductive,velivckovic2017graph}. Note that for the first layer, the input node embedding $\vec{h}^0_v$ is typically initialized with the node's features from $\vec{X}_V$. For brevity, we denote the output node representations from the last layer as $\vec{h}_v$.

Finally, the whole-graph representation can be obtained through a readout operation:
\begin{align}
    \vec{h}_G = \textsc{ReadOut}(\vec{h}_v : v \in V),
\end{align}
which aggregates the node representations across the entire graph, such as sum pooling or mean pooling.

\section{Few-Shot Learning Problems on Graphs}\label{sec.pro}

\begin{figure}[t]
\centering
\includegraphics[width=1\linewidth]{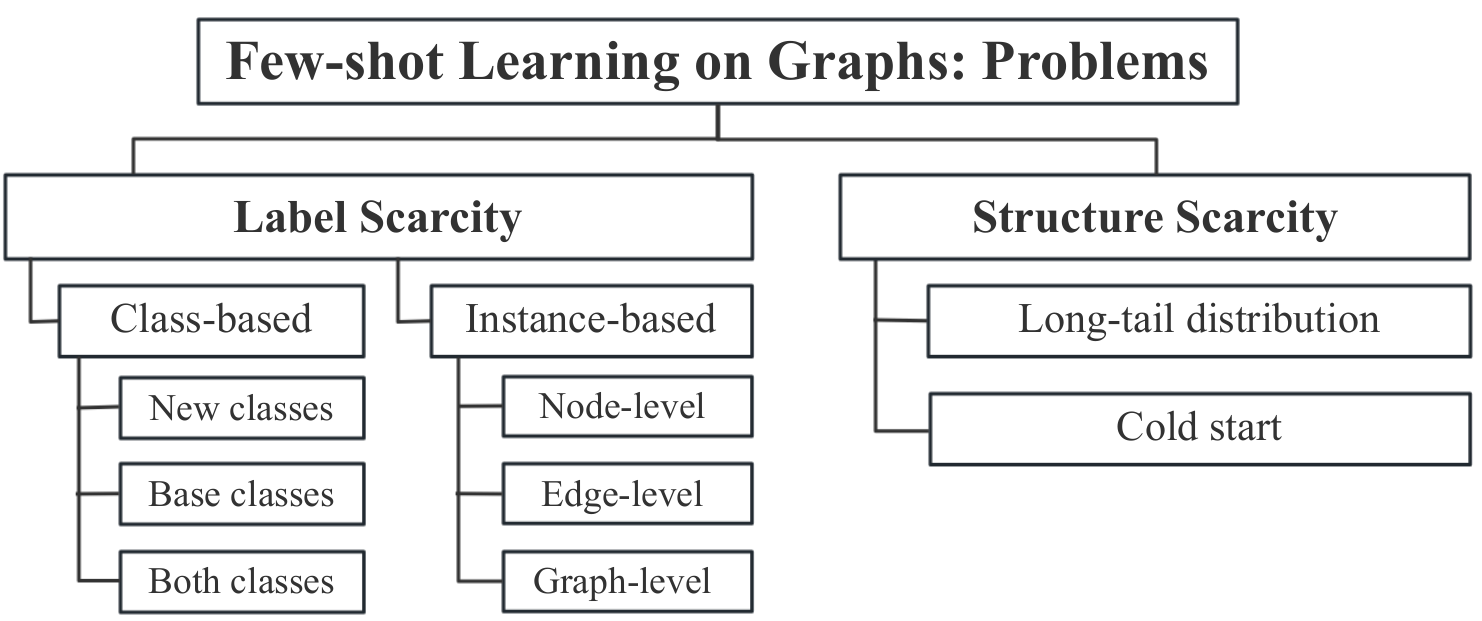} 
\caption{Taxonomy of few-shot learning problems on graphs.}
\label{fig.taxonomy-problem}
\end{figure}

Few-shot learning on graphs has gained significant research interest due to the frequent data scarcity issues in real-world graphs. Based on the type of data scarcity, we categorize few-shot learning problems on graphs into two groups: \emph{label scarcity} and \emph{structure scarcity}. On one hand, similar to the challenges faced in NLP and CV \cite{wang2020generalizing,song2023comprehensive,wang2019few}, 
label scarcity, or the lack of labeled data, remains an important challenge in few-shot learning on graphs.
On the other hand, unlike text and image data, graphs possess a non-euclidean topological structure. As a result, structure scarcity in graphs emerges as a second challenge that could adversely impact the learning of effective representations.
Researchers have studied various problems related to the two challenges, and in this section, we aim to categorize the literature based on the taxonomy illustrated in Fig.~\ref{fig.taxonomy-problem}.

\subsection{Label Scarcity Problem on Graphs}\label{sec.label-scarcity}
As acquiring labels is usually difficult or costly, label scarcity is a common problem in real-world applications. However, the performance of supervised methods heavily relies on a large amount of labeled data as supervision. Consequently, traditional supervised methods perform poorly when provided with limited labeled data, motivating few-shot learning approaches to address the label scarcity problem on graphs. 
We further categorize the label scarcity problem into \textit{class-based} and \textit{instance-based} label scarcity based on their respective class settings and target instances.


\subsubsection{Class-based Label Scarcity}
Let \( C \) denote the entire set of classes on a graph, which consists of two subsets: the base class set \( C_{\text{base}} \) for model training, and the new class (also called novel or unseen class) set \( C_{\text{new}} \) for testing, such that \( C = C_{\text{base}} \cup C_{\text{new}} \) and \( C_{\text{base}} \cap C_{\text{new}} = \emptyset \). 
Note that label scarcity could happen in either subsets or both, as follows.

\stitle{Label scarcity in new classes.}
In this setting, the goal is to learn a prior using sufficient labeled data from the base classes \( C_{\text{base}} \), and then transfer the prior knowledge to a novel task involving the new classes \( C_{\text{new}} \). The labeled data for the new classes form the \emph{support set}, while the unlabeled data from these new classes constitute the \emph{query set}. When the support set contains exactly \( K \) labeled samples for each of the \( N \) classes from \( C_{\text{new}} \), this setup is called an \( N \)-way \( K \)-shot problem. In particular, a few-shot problem refers to the scenario where $K$ is a small value. To address the few-shot problem for the new classes, researchers have developed meta-learning based approaches (see Sect.~\ref{sec.meta-learning}) to learn prior knowledge from labeled data in the base classes, and further transfer and adapt this knowledge to predict labels in the new classes \cite{zhou2019meta,wang2020graph,huang2020graph,yao2020graph,zhang2022mul,wang2022task,zhou2022task,liu2022few,wen2023augmenting,sun2023all,sheng2020adaptive}. However, the effectiveness of these approaches still heavily depends on the availability of abundant labeled data in a large number of base classes, which might also be challenging to obtain in real-world applications. 

\stitle{Label scarcity in base classes.}
This setting occurs when there is limited labeled data in \( C_{\text{base}} \), resulting in insufficient prior knowledge to transfer and adapt to the new classes. More commonly, when there is no labeled data for the base classes or no base classes at all, meta-learning approaches become inapplicable. Therefore, researchers have turned to self-supervised methods to pre-train graph encoders  without requiring any labeled data or base classes \cite{kipf2016variational,velivckovic2018deep,sun2019infograph,jiao2020sub,hu2020gpt,zhu2020deep,qiu2020gcc,you2020graph,xu2021self,xia2022simgrace,bei2023cpdg,sun2022self,li2023s,hou2023graphmae2,zhang2023structure} (see Sect.~\ref{sec.pre-training-strategy}). The pre-trained model is subsequently adapted to novel downstream tasks involving new classes through fine-tuning the pre-trained weights (see Sect.~\ref{sec:pre-train:fine-tune}). The fine-tuning process typically requires a reasonable amount of labeled data and is usually not considered as a few-shot problem.

\stitle{Label scarcity in both classes.}
This is a combination of the two settings above, where labeled data are limited in both \( C_{\text{base}} \) and  \( C_{\text{new}} \). Similar to the label scarcity in base classes, 
self-supervised methods are often leveraged to first pre-train a graph encoder. However, with limited labeled data in the new classes as well, vanilla fine-tuning can be challenging in the few-shot context. Specifically, vanilla fine-tuning updates the full parameter set of the pre-trained model, which is not only expensive for large models but also prone to overfitting to the few-shot support set. Therefore, parameter-efficient adaptation methods \cite{sun2022gppt,liu2023graphprompt,yu2023generalized,li2023adaptergnn,gui2023g,liu2023molca} are more suitable for these few-shot learning problems on graphs (see Sect.~\ref{sec:pre-train:param-efficient}).



\subsubsection{Instance-based Label Scarcity}
An orthogonal categorization of label scarcity problems on graphs can be defined by the affected instances at \emph{node}, \emph{edge} or \emph{graph} levels.
We summarize the literature and their applications w.r.t.~the instance-based categorization in Table~\ref{table.application-label-scarcity}.

\begin{table}[tbp] 
    \centering
    \small
    \addtolength{\tabcolsep}{-1mm}
    \caption{Summary of instance-based label scarcity learning on graphs and their applications.}
    \label{table.application-label-scarcity}%
    \resizebox{1\linewidth}{!}{
    \begin{tabular}{@{}c|c|c@{}}
    \toprule
    \makecell{\textbf{Instance}} & \textbf{Application domain} & \textbf{Literature}\\\midrule
    \multirow{5}{*}{Node} 
    & Academic network & \makecell{\cite{zhou2019meta,wang2020graph,huang2020graph,yao2020graph,zhang2022mul,wang2022task,zhou2022task,liu2022few,wen2023augmenting}\\ \cite{wen2023prompt,yu2023hgprompt,yu2023multigprompt,sun2022gppt,ge2023enhancing,chen2023ultra,tan2023virtual,sun2023all,wang2023contrastive}}\\
    & Social network & 
    \cite{zhou2019meta,wang2020graph,yao2020graph,liu2021relative,zhou2022task,liu2022few,
    sun2022gppt,zhu2023sgl,sun2023all}
    \\
    & E-commerce network & \cite{wang2020graph,liu2021relative,wang2022task,liu2022few,sun2023all,sun2022gppt} \\
    & Protein-protein interaction & \cite{wang2022graph}\\
    & Traffic flow & \cite{lu2022spatio} \\\midrule
    \multirow{4}{*}{Edge} 
    & Drug-drug interaction & \cite{baek2020learning}\\
    & Protein multimer structure & \cite{gao2024protein}\\
    & E-comm./academic network & \cite{zhu2023few}\\
    & Knowledge graphs & \cite{xiong2018one,chen2019meta,zhang2020few,wang2021reform,niu2021relational,jambor2021exploring,zhang2022adapting,luo2023normalizing,wu2023hierarchical}
    \\\midrule
    \multirow{3}{*}{Graph} 
    & Molecular graph & \cite{wang2022molecular,zaidi2022pre,wang2023automated,liu2022pre,li2023knowledge,zhu2023dual,zhu2022unified}\\
    & Protein graph  & \cite{guo2022self,zhang2023protein,gao2024protein}\\
    & Social network & \cite{chauhan2019few,sun2023all}\\
    \bottomrule
    \end{tabular}}
\end{table}

\stitle{Node-level label scarcity.}
Nodes are fundamental components of graphs, making node-level tasks, such as node classification, highly prevalent.
The lack of node labels leads to node-level label scarcity, where only a limited number of labeled nodes are available per class on the graph.
In response to this challenge, a range of methods have been developed to address 
various node-centric applications. For example, in academic networks, where many authors or papers have unknown fields, researchers propose various strategies to classify authors or papers into their primary field, topic or venues with limited labels \cite{zhou2019meta,wang2020graph,huang2020graph,yao2020graph,zhang2022mul,wang2022task,zhou2022task,liu2022few,wen2023augmenting,wen2023prompt,yu2023hgprompt,yu2023multigprompt,sun2022gppt,chen2023ultra,tan2023virtual,sun2023all,wang2023contrastive,yan2024inductive,ma2024hetgpt,ge2023enhancing}. 
In the realm of social networks, where interests of users or categories of organizations may be difficult or costly to obtain, previous studies \cite{zhou2019meta,wang2020graph,yao2020graph,liu2021relative,zhou2022task,liu2022few,sun2022gppt,zhu2023sgl,sun2023all,yan2024inductive} propose effective solutions to overcome these obstacles. Moreover, with the growing popularity of e-commerce platforms, some recent research \cite{wang2020graph,liu2021relative,wang2022task,liu2022few,sun2023all,sun2022gppt,yan2024inductive} tackles product categorization in e-commerce networks.
Additionally, recent studies also explore node-level label scarcity in protein property prediction \cite{wang2022graph} and traffic speed prediction \cite{lu2022spatio}. In a more challenging scenario, some studies address zero-shot node-level learning tasks in academic or e-commerce networks, where no labeled node is available at all \cite{wang2021zero,yue2022dual,wen2023augmenting,wen2023prompt}.

\stitle{Edge-level label scarcity.}
Edges serve as explicit connectors between nodes in a graph. In knowledge graphs, various kinds of  relation exist between entities. Hence, while nodes represent entities, edges are associated with labels that represent semantic relationships between entities \cite{toutanova2015representing,xiong2017deeppath}. Likewise, in a drug-drug interaction network, edge labels indicate the nature of interaction between two drugs \cite{leskovec2016snap,ryu2018deep}.
However, in real-world applications, such labeled edges are often scarce: In knowledge graphs, it is impractical to exhaustively annotate all relationships, while in drug interactions, expensive laboratory setups may be needed. The edge-level scarcity motivates researchers to investigate solutions for learning with few-shot edge labels.
In knowledge graphs, many works aim to predict the semantic relationship between entities \cite{xiong2018one,chen2019meta,zhang2020few,wang2021reform,niu2021relational,jambor2021exploring,zhang2022adapting,luo2023normalizing,wu2023hierarchical}, when novel relations are introduced with only one or a few labeled example available.
In biochemical applications, few-shot solutions for drug interaction prediction \cite{baek2020learning} and protein multimer structure prediction \cite{gao2024protein} have been proposed. In e-commerce and academic networks, product-sharing prediction and co-authorship prediction have been explored in few-shot settings \cite{zhu2023few}. 

\stitle{Graph-level label scarcity.}
Beyond predictions on individual nodes or edges, numerous important tasks are performed at the graph level, such as predicting properties or categories for subgraphs or whole graphs. Similar to the challenge at the node or edge level, we often face a shortage of labeled subgraph or graph samples. 
For example, several studies \cite{chauhan2019few,sun2023all} propose solutions for categorizing user communities in social networks. Moreover, few-shot approaches have been devised for property prediction for molecules \cite{wang2022molecular,zaidi2022pre,wang2023automated,liu2022pre,li2023knowledge,zhu2023dual,zhu2022unified} and proteins \cite{guo2022self,zhang2023protein,gao2024protein}, where obtaining labels, such as solubility and permeability, often requires significant resources.

\subsection{Structure Scarcity Learning on Graphs}\label{sec.structure-scarcity}
In contrast to label scarcity, which stems from the high cost or difficulty of obtaining annotations, 
structure scarcity arises from the inherent sparsity in graph topology, where many nodes are incident to very few edges. Since graph representation learning fundamentally relies on the richness of topological structures between nodes, structure scarcity presents a significant hurdle for effective learning. In the real world, structure scarcity in graphs is a common issue, which has attracted increasing research interest. We categorize the literature on structure scarcity into two subtypes: \emph{long-tailed distribution} and \emph{cold-start} learning problems, based on their goal and application scenarios 
as summarized in Table~\ref{table.application-structure-scarcity}. 
Specifically, the long-tailed problem addresses the challenge of learning from an imbalanced distribution, where a large number of nodes have few connections, while the cold-start problem focuses on learning representations for new nodes with no or very few connections.

\begin{table}[tbp] 
    \centering
    \small
    \caption{Summary of structure scarcity learning on graphs and their applications.}
    \label{table.application-structure-scarcity}%
    \resizebox{1\linewidth}{!}{
    \begin{tabular}{@{}c|c|c@{}}
    \toprule
    \textbf{Goal}
    & \textbf{Application domain} & \textbf{Literature}\\\midrule
    \multirow{5}{*}{\makecell{Long-tailed\\distribution}} 
    & Academic network & \cite{kojaku2021residual2vec,tang2020investigating,yun2022lte4g,virinchi2023blade,liu2023generalized,ju2024graphpatcher,liang2023tackling}\\
    & Social network & \cite{kojaku2021residual2vec, wu2019net, tang2020investigating, xia2022cengcn, liu2020towards, liu2021tail, virinchi2023blade,liu2023generalized,ju2024graphpatcher}
    \\
    & E-commerce network & \cite{liu2021tail, virinchi2023blade, niu2020dual,ju2024graphpatcher,liang2023tackling} \\
    & Protein-protein interaction & \cite{kojaku2021residual2vec}\\
    & Air traffic control & \cite{wu2019net} \\\midrule
    \multirow{2}{*}{Cold start} 
    & Social network & \cite{lee2019melu,pan2019warm,yang2022few, lu2020meta, hao2021pre, kim2024content}\\
    & E-commerce network & \cite{lee2019melu,lu2020meta, zheng2021cold}\\
    \bottomrule
    \end{tabular}}
\end{table}

\stitle{Long-tailed distribution.}
Many datasets exhibit long-tailed distributions \cite{zhang2023deep}, where a large proportion of occurrences are concentrated in the low-frequency region, or the ``tail''. In the context of graphs, only a small number of nodes, referred to as \emph{head} nodes, are incident to a large number of edges (\ie, high-degree), while the majority of nodes, termed \emph{tail} nodes, have very few edges (\ie, low-degree) \cite{liu2020towards}. Thus, tail nodes are surrounded by a small neighborhood with a scarcity of neighbor information, limiting effective learning of their representations. 
Formally, given a graph $G=(V,E)$, the set of head nodes is denoted by $V_\text{head}$, and the set of tail nodes by $V_\text{tail}$, such that $V_\text{head} \cup V_\text{tail}=V$ and $V_\text{head} \cap V_\text{tail}=\emptyset$. 
The separation of the two subsets is often determined by a predefined threshold on the node degree \cite{tang2020investigating,liu2020towards}. 


Long-tailed distributions are common in real-world networks. For example, in academic networks, a small number of authors are highly prolific and linked to many papers, while the majority have only a few connections to papers \cite{kojaku2021residual2vec,tang2020investigating,yun2022lte4g,virinchi2023blade,ju2024graphpatcher,liang2023tackling}. In social networks, a few celebrities have a vast number of followers, while most users have far fewer \cite{kojaku2021residual2vec, wu2019net, tang2020investigating, xia2022cengcn, liu2020towards, liu2021tail,  virinchi2023blade,liu2023generalized,ju2024graphpatcher}. In e-commerce and recommendation systems \cite{liu2021tail, virinchi2023blade, niu2020dual,ju2024graphpatcher,liang2023tackling}, the disparity in the number of connections across users or items is also evident. The challenge of long-tailed distributions has also been studied in other graph-centric applications, ranging from protein-protein interaction \cite{kojaku2021residual2vec} to air traffic control \cite{wu2019net}.

\stitle{Cold-start learning.}
The cold-start problem in graphs arises when making predictions for new nodes, which often have no or very few existing edges. 
Conventionally, models are trained on existing graph structures where nodes have sufficient connections to other nodes, leading to poor performance for new nodes with few connections. 
This problem is prevalent in social networks and e-commerce systems, where new nodes are continually added.
Specifically, in social networks, new users often begin with few interactions with existing users or content, motivating various solutions to the cold-start problem \cite{lee2019melu,pan2019warm,yang2022few, lu2020meta, hao2021pre, kim2024content}. Similarly, on commerce platforms, new users or items often lack interactions with other users or items during their initial phase \cite{lee2019melu,lu2020meta,zheng2021cold}.

\begin{figure}[t]
\centering
\includegraphics[width=1\linewidth]{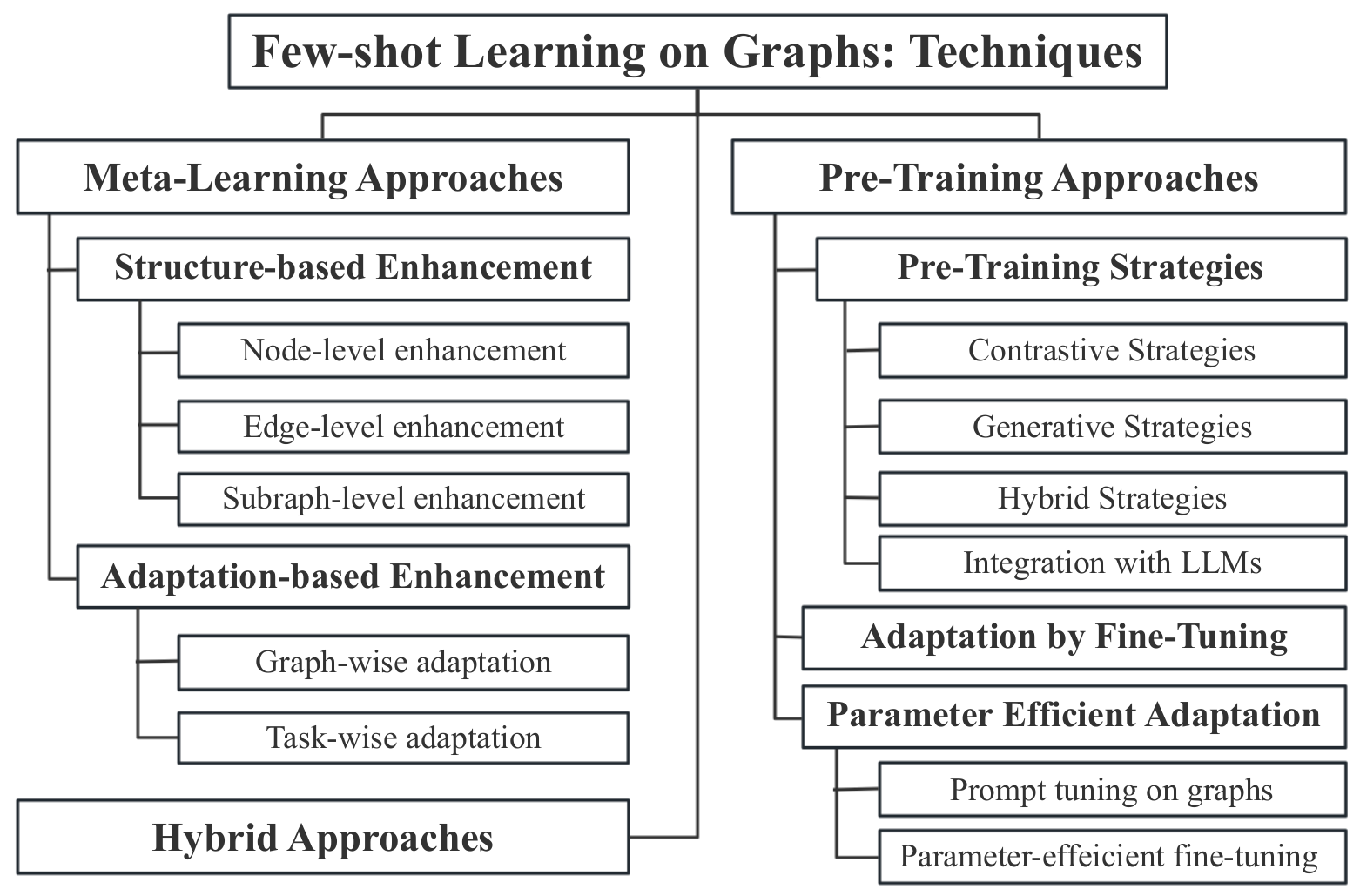} 
\caption{Taxonomy of few-shot learning techniques on graphs.}
\label{fig.taxonomy-technique}
\end{figure}

\section{Meta-Learning Techniques on Graphs}\label{sec.meta-learning}
Meta-learning is an important family of few-shot learning techniques, represented in the first branch of our taxonomy in Fig.~\ref{fig.taxonomy-technique}.
Prevailing episodic meta-learning methods \cite{vinyals2016matching,snell2017prototypical,finn2017model} are designed to learn a prior from base classes, which is transferable to new classes in downstream tasks. These methods typically assume an abundance of labeled data across a large number of base classes, while the downstream new classes have few labeled samples, \ie, the setting of label scarcity in new classes as discussed in Sect.~\ref{sec.label-scarcity}.

In this section, we first review standard meta-learning techniques. We then categorize and discuss literature that enhances the standard methods for few-shot learning on graphs.

\subsection{Standard Meta-learning Techniques}
\begin{figure}[t]
\centering
\includegraphics[width=1\linewidth]{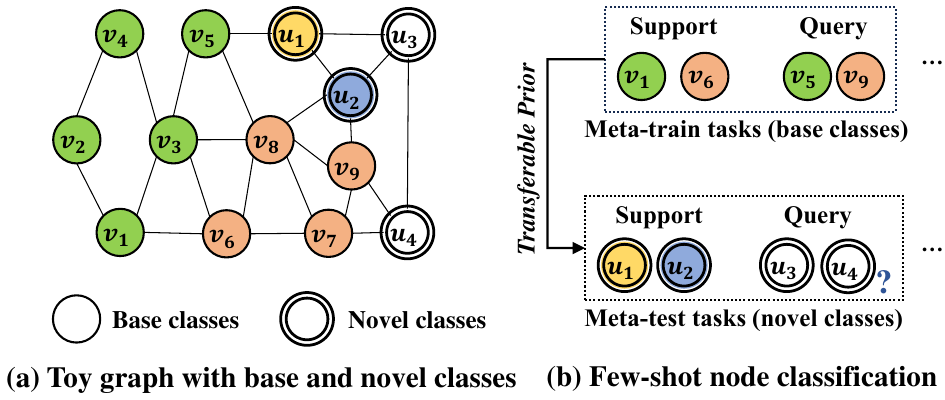}
\vspace{-2mm}
\caption{An illustration of meta-learning on graph.}
\label{fig.meta-learning}
\end{figure}

In standard episodic meta-learning methods, such as Matching Networks (MN) \cite{vinyals2016matching}, Model-Agnostic Meta-Learning (MAML) \cite{finn2017model} and Prototpyical Networks (Protonets) \cite{snell2017prototypical}, we first construct a series of \emph{meta-training tasks} from the base classes to learn prior knowledge, which is then adapted to  downstream few-shot \emph{meta-testing tasks} involving the new classes. 
The concept is illustrated in Fig.~\ref{fig.meta-learning}, using few-shot node classification on a graph as an example. In this scenario, meta-training tasks are constructed using base classes with ample labels, whereas a meta-testing task involves new classes with only few-shot labels. Each task comprises a \emph{support set} and a \emph{query set}, which function as training and testing data, respectively, within each task \cite{finn2017model}. The support set is always labeled, whereas the query set is only labeled in meta-training but is unlabeled for inference and prediction in meta-testing. The prior is optimized on the meta-training tasks to enable rapid, lightweight adaptation using the support set that can minimize the loss on the query set. Hence, for the prior to be transferable to meta-testing tasks, such meta-learners assume that the meta-training and -testing tasks are sampled from an i.i.d.~task distribution. 

Formally, in the \emph{meta-training} phase, consider a set of meta-training tasks $\bT_{\text{train}} = \{\bT^1_{\text{train}}, \bT^2_{\text{train}}, \ldots, \bT^n_{\text{train}}\}$. Each task $\bT^i_{\text{train}} = (\bS^i_{\text{train}}, \bQ^i_{\text{train}}) \in \bT_{\text{train}}$ consists of a support set $\bS^i_{\text{train}}$ and a query set $\bQ^i_{\text{train}}$. Subsequently, the process of meta-learning, also called \emph{learning-to-learn} \cite{finn2017model}, involves optimizing the loss over the set of meta-training tasks $\bT^i_{\text{train}}$, as follows.
\begin{equation}
    \omega^* = \arg\min_{\omega} \mathbb{E}_{\bT^i_{\text{train}} \in \bT_{\text{train}}} \bL(\bT^i_{\text{train}}; \omega),
\end{equation}
where $\omega^*$ represents the prior or meta-knowledge extracted by the meta-learner from the meta-training tasks, describing ``how to learn''. The meta-knowledge $\omega^*$ can then be transferred to downstream novel tasks, \ie, training a model for the downstream task under the guidance of $\omega^*$. Specifically, in the \emph{meta-testing} phase, consider a set of \emph{meta-testing} tasks $\bT_{\text{test}} = \{\bT^1_{\text{test}}, \bT^2_{\text{test}}, \ldots, \bT^m_{\text{test}}\}$, where each task $\bT^i_{\text{test}} = (\bS^i_{\text{test}}, \bQ^i_{\text{test}}) \in \bT_{\text{test}}$ also includes a few-shot support set $\bS^i_{\text{test}}$ and an unlabeled query set $\bQ^i_{\text{test}}$. Given the extracted meta-knowledge $\omega^*$ and a meta-testing task $\bT^i_{\text{test}} = (\bS^i_{\text{test}}, \bQ^i_{\text{test}}) \in \bT_{\text{test}}$, meta-learning trains a specialized downstream model for task $\bT^i_{\text{test}}$ by optimizing the loss on its support set $\bS^i_{\text{test}}$:
\begin{equation} \label{eq.test-support}
    \theta^*_i = \arg\min_{\theta} \bL(\bS^i_{\text{test}}, \omega^*; \theta),
\end{equation}
where $\theta$ denotes the parameters of the downstream model. The optimized parameters $\theta^*_i$ can then be used for prediction on the query set $\bQ^i_{\text{test}}$ of the task $\bT^i_{\text{test}}$. 

Toward few-shot learning on graphs, these standard meta-learning methods can be enhanced in various ways. We categorize the literature based on their graph-specific enhancements within the meta-learning framework into two types: \emph{structure-based} enhancements and \emph{adaptation-based} enhancements, as summarized in Table~\ref{table.meta-learning-structure-based} and Table~\ref{table.meta-learning-adaptation-based}, respectively.

\subsection{Structure-based Enhancement on Graphs}
A unique characteristic of graph data lies in the connectivity  structures between nodes, 
presenting crucial information for coping with graph data. 
Consequently, many meta-learning methods for graph-based tasks have focused on exploiting graph structures to enhance the learning of the prior.
Based on the type of graph structures, they can be further grouped into node-, edge-, and subgraph-level enhancements.  

\stitle{Node-level enhancement.}
The nodes of a graph encompass implicit structural information for enhancing the meta-learning model. 
Recent studies predominantly focus on differentiating node weights in a task to reflect their varying structural importance. 
Specifically, in Protonets-based methods such as GPN \cite{ding2020graph} and FAAN \cite{sheng2020adaptive}, it is observed that each support node contributes variably to the prototype of each class in a task. Thus, they 
propose to evaluate the importance of nodes in the support set and adjust their contribution \wrt the query node when constructing the prototypes.

\stitle{Edge-level enhancement.}
Edges are fundamental to graphs as they represent explicit first-order structures. Moreover, a sequence of edges forms a path, which is capable of representing more complex high-order interactions among distant nodes. As a result, many meta-learning methods on graphs incorporate edge- or path-based enhancements. On one hand, edge-based enhancement leverages auxiliary information associated with edges, thereby extending beyond mere node-level information. For instance, to boost few-shot link prediction in knowledge graphs, HMNet \cite{xiao2021hmnet} adopts matching networks (MN) as its meta-learning framework, performing matching for both entities and relations. In particular, for relation-level matching, it calculates a relation-aware score using a relation encoder to enrich the link prediction task. 
Another work \cite{wang2019tackling} develops a generative model to generate new triplets, serving to augment the sparse links within a MAML-based framework. 
On the other hand, paths can capture long-range dependencies between distant nodes.
RALE \cite{liu2021relative} proposes to exploit the paths between each query node and the support nodes for task-level dependencies, and between each target node and the hub nodes for graph-level dependencies.
Furthermore, MetaHIN \cite{lu2020meta} employs meta-paths \cite{sun2011pathsim} to explore heterogeneous semantic relationships between users and items, aiming to acquire finer-grained semantic priors for new tasks.

\stitle{Subgraph-level enhancement.}
Subgraphs provide rich contextual information to its containing nodes. Hence, recent studies have utilized subgraphs to enhance the expressiveness of nodes.
G-Meta \cite{huang2020graph} generates class prototypes based on the contextual subgraph of each support node, and similarly expands each query node to its subgraph, to enhance the robustness of prototype and query representations.
Similarly, GEN \cite{baek2020learning} proposes a graph extrapolation network to extrapolate knowledge through the neighbors of the support set, which essentially exploits the one-hop subgraphs of the support nodes. 
Moreover, meta-tail2vec \cite{liu2020towards} introduces the notion of locality-aware tasks, where the support set is sampled from the neighborhood subgraph of the query node, leveraging the structural proximity between support and query nodes.   

\begin{table}[tbp] 
    \centering
    \small
    \addtolength{\tabcolsep}{-1mm}
    \caption{Structure-based meta-learning enhancement for few-shot learning on graphs.\label{table.meta-learning-structure-based}}
    \resizebox{1\linewidth}{!}{
    \begin{tabular}{@{}c|c|c|ccc@{}}
    \toprule
    \multirow{2}*{\textbf{Method}} & {\textbf{Structure}} & \multirow{2}*{\textbf{Meta-learner}} & \multicolumn{3}{c}{\textbf{Task}}\\
    &\textbf{enhancement} & & Node & Edge & Graph\\\midrule
    GPN \cite{ding2020graph}
    & node
    & Protonets
    & $\checkmark$
    & $\times$
    & $\times$
    \\
    FAAN \cite{sheng2020adaptive}
    & node
    & Protonets
    & $\times$
    & $\checkmark$
    & $\times$
    \\
    HMNet \cite{xiao2021hmnet}
    & edge
    & MN
    & $\times$
    & $\checkmark$
    & $\times$
    \\
    \cite{wang2019tackling}
    & edge
    & MAML
    & $\times$
    & $\checkmark$
    & $\times$
    \\
    RALE \cite{liu2021relative}
    & path
    & MAML
    & $\checkmark$
    & $\times$
    & $\times$
    \\
    MetaHIN \cite{lu2020meta}
    & path
    & MAML
    & $\checkmark$
    & $\times$
    & $\times$
    \\
    G-Meta \cite{huang2020graph}
    & subgraph
    & hybrid
    & $\checkmark$
    & $\checkmark$
    & $\times$
    \\
    GEN \cite{baek2020learning}
    & subgraph
    & Protonets
    & $\times$
    & $\checkmark$
    & $\times$
    \\
    meta-tail2vec \cite{liu2020towards}
    & subgraph
    & MAML
    & $\checkmark$
    & $\checkmark$
    & $\times$
    \\
    \bottomrule
    \end{tabular}}
\end{table}

\subsection{Adaptation-based Enhancement on Graphs}
Achieving rapid adaptation to the query set through the support set is a key requirement of meta-learning. 
In particular, adaptation enables the customization of a globally shared prior into a localized or specialized model for each task, thereby not only preserving the unique traits of each individual task but also capitalizing on shared common knowledge.
Hence, many approaches aim to enhance the adaptation mechanism for few-shot learning on graphs.

\stitle{Graph-wise adaptation.} GFL \cite{yao2020graph} recognizes the topological variances across different graphs, and thus customizes a global prior for each individual graph. More concretely, it generates class prototypes specifically tailored to each graph, utilizing a gate function based on graph-specific representation that is applied to the global prior.
MI-GNN \cite{wen2021metainductive} adopts a similar idea and employs a Feature-wise Linear Modulation (FiLM) \cite{perez2018film} to modulate the global prior for each graph before undergoing a further task-wise adaptation within a MAML framework. 

\stitle{Task-wise adaptation.}
Meanwhile, there is potential for enhancement in task-wise adaptation. In the context of multi-label few-shot classification, MetaTNE \cite{lan2020node} observes an increase in task diversity due to the multi-label setting. Thus, adopting a uniform node embedding across such diverse tasks may result in inadequate expressiveness, given that the same node could be associated with different labels in different tasks. To address this, MetaTNE conducts an additional adaptation for the node embeddings \wrt the query set in each task, thereby better adapting to the characteristics of the task.
Moreover, variation in feature distribution across tasks can significantly impair the performance of meta-learning. To mitigate this issue, AMM-GNN \cite{wang2020graph} capitalizes on the FiLM mechanism \cite{perez2018film} to modulate the feature matrix for each task. It customized a task-specific feature matrix, enabling a globally shared GNN model to apply these tailored features for task-wise adaptation in few-shot node classification.

\stitle{Others.} In addition to enhancing graph- or task-wise adaptation, several approaches enhances downstream adaptation in various other ways. AS-MAML \cite{ma2020adaptive} improves adaptation from an optimization standpoint. It employs a reinforcement learning-based controller to determine the optimal step size for the adaptation process in MAML. MetaDyGNN \cite{yang2022few} introduces hybrid adaptations for dynamic graphs that entails both time- and node-wise adaptations within a MAML framework to solve few-shot link prediction problems.

\begin{table}[tbp] 
    \centering
    \small
    \addtolength{\tabcolsep}{-1mm}
    \caption{Adaptation-based meta-learning enhancement for few-shot learning on graphs.\label{table.meta-learning-adaptation-based}}
    \resizebox{1\linewidth}{!}{
    \begin{tabular}{@{}c|c|c|ccc@{}}
    \toprule
    \multirow{2}*{\textbf{Method}} & {\textbf{Adaptation}} & {\textbf{Meta}} & \multicolumn{3}{c}{\textbf{Task}}\\
    &\textbf{enhancement} &\textbf{learner} & Node & Edge & Graph\\\midrule
    GFL \cite{yao2020graph}
    & graph
    & MAML
    & $\checkmark$
    & $\times$
    & $\times$
    \\    
    MI-GNN \cite{wen2021metainductive}
    & graph
    & hybrid
    & $\checkmark$
    & $\times$
    & $\times$
    \\
    MetaTNE \cite{snell2017prototypical}
    & task
    & Protonets
    & $\checkmark$
    & $\times$
    & $\times$
    \\
    AMM-GNN \cite{wang2020graph}
    & task
    & MAML
    & $\checkmark$
    & $\times$
    & $\times$
    \\
    AS-MAML \cite{ma2020adaptive}
    & step size
    & MAML
    & $\times$
    & $\times$
    & $\checkmark$
    \\
    MetaDyGNN \cite{yang2022few}
    & hybrid
    & MAML
    & $\times$
    & $\checkmark$    
    & $\times$
    \\
    \bottomrule
    \end{tabular}}
\end{table}

\subsection{Discussion}
In summary, existing research for few-shot learning on graphs often enhances a standard meta-learner, either through structural augmentation or by refining the adaptation process.
However, these approaches share a common drawback inherent to meta-learning approaches. First, they still require abundant labels for a base set during the meta-training phase, which can be challenging or costly to acquire.
Second, they fail to leverage the vast amount of unlabeled data to learn a more comprehensive prior. 
Third, they are limited by the i.i.d.~assumption in task distribution, and cannot handle different types of tasks downstream. 
This leads to a further challenge: Can we address a \emph{diverse} range of few-shot tasks on graphs \emph{without an extensively annotated base set}, while \emph{utilizing abundant unlabeled graphs}? Recently, pre-training on graphs has emerged as a promising solution to this challenge.

\section{Pre-training Techniques on Graphs}\label{sec.pre-training}
\begin{figure*}[t]
    \centering
    \includegraphics[width=0.95\linewidth]{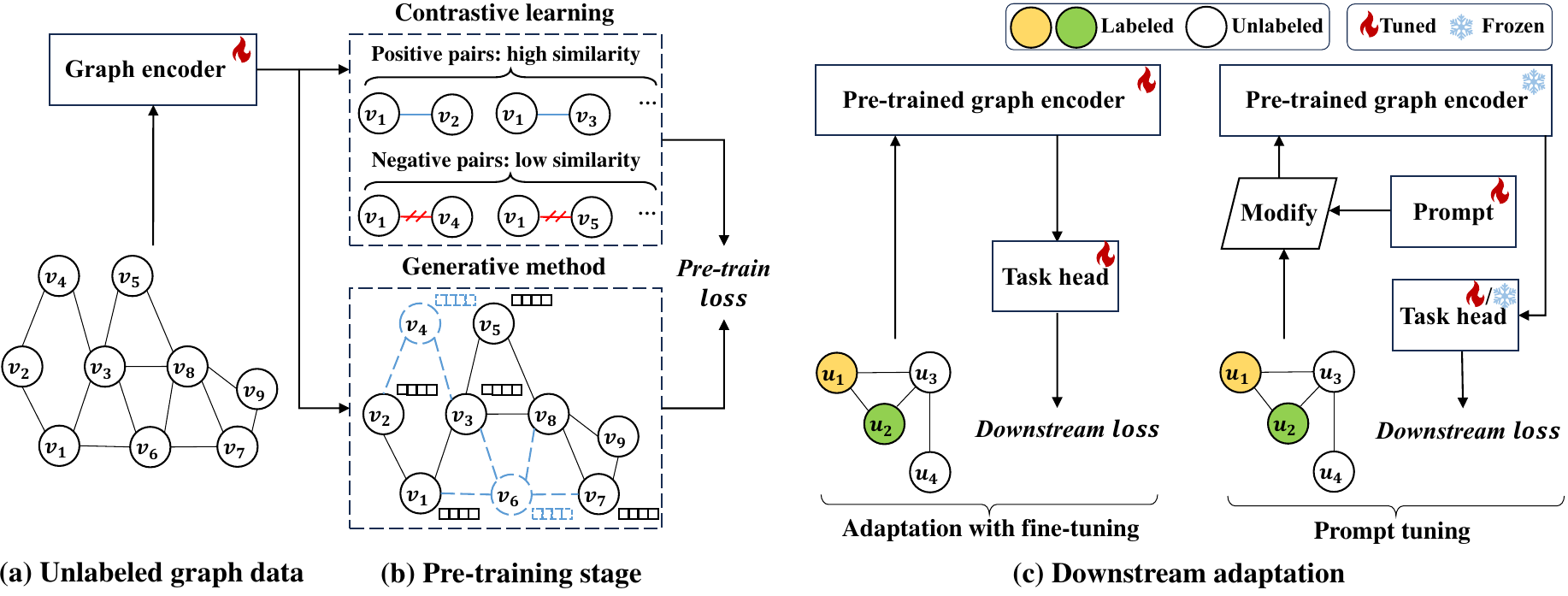}
    \caption{An illustration of pre-training and adaptation for few-shot learning on graphs.}
    \label{fig. pre-training}
\end{figure*}

Given the abundance of unlabeled data in many domains and the flexibility to deal with many types of downstream tasks, pre-training approaches have been popularized \cite{kenton2019bert}. 
In graph context, the \emph{pre-training stage} typically utilizes self-supervised methods on unlabeled graph data to train a graph encoder, which aims to capture task-agnostic intrinsic properties of graphs, such as node features and local or global structures. Subsequently, the pre-trained graph encoder, serving as a form of prior knowledge, can be used to address a variety of downstream tasks through an \emph{adaptation} stage. An overview of the pre-training and adaptation process is depicted in Fig.~\ref{fig. pre-training}.

In this section, we provide a taxonomy of existing work in this direction, covering both pre-training and   adaptation.

\begin{table}[tbp] 
    \centering
    \small
    \addtolength{\tabcolsep}{-1mm}
    \caption{Summary of contrastive graph pre-training.}\label{table.contrastive}
    \resizebox{1\linewidth}{!}{
    \begin{tabular}{@{}c|c|c|c@{}}
    \toprule
    {\textbf{Method}} &{\textbf{Instance}} & {\textbf{Augmentation}} & {\textbf{Graph types}}\\
    \midrule
    GRACE \cite{zhu2020deep}
    & node
    & uniform
    & general
    \\
    GCC \cite{qiu2020gcc}
    & graph
    & uniform
    & general
    \\
    GraphCL \cite{you2020graph}
    & graph
    & uniform
    & general
    \\
     SimGRACE \cite{xia2022simgrace}
    & graph
    & perturbing encoder
    & general
    \\
    GraphLoG \cite{xu2021self}
    & dataset
    & uniform
    & general
    \\
    DGI \cite{velivckovic2018deep}
    & cross-scale
    & uniform
    & general
    \\
    InfoGraph \cite{sun2019infograph}
    & cross-scale
    & uniform
    & general
    \\
    Subg-Con \cite{jiao2020sub}
    & cross-scale
    & uniform
    & general
    \\
    MVGRL \cite{hassani2020contrastive}
    & cross-scale
    & diffusion
    & general
    \\
    JOAO \cite{you2021graph}
    & graph
    & adaptive to loss
    & general
    \\
    GCGM \cite{ijcai2024gcgm}
    & node
    & adaptive to loss
    & general
    \\
    You \etal~\cite{you2022bringing}
    & graph
    & view generator
    & general
    \\
    GCA \cite{zhu2021graph}
    & node
    & adaptive to instance
    & general
    \\
    HeCo \cite{wang2021self}
    & node
    & uniform
    & hetero.
    \\
    CPT-HG \cite{jiang2021contrastive}
    & cross-scale
    & uniform
    & hetero.
    \\
    PT-HGNN \cite{jiang2021pre}
    & cross-scale
    & uniform
    & hetero.
    \\    
    SelfRGNN \cite{sun2022self}
    & node
    & curvature over time
    & dynamic
    \\
    DDGCL \cite{tian2021self}
    & graph
    & uniform
    & dynamic
    \\
    CPDG \cite{bei2023cpdg}
    & cross-scale
    & temporal-aware sampling
    & dynamic
    \\
    GearNet \cite{zhang2022protein}
    & graph
    & uniform
    & 3D
    \\    
    \bottomrule
    \end{tabular}}
\end{table}

\subsection{Strategies for the Pre-training Stage}\label{sec.pre-training-strategy}
Current self-supervised tasks (also known as \emph{pretext} tasks) for graph pre-training predominantly fall into two broad categories: \emph{contrastive} and \emph{generative} strategies, although many approaches employ both strategies simultaneously to extract more comprehensive knowledge from graphs. More recently, some graph pre-training approaches have begun to leverage the power of pre-trained large language models.

\subsubsection{Contrastive Strategies}
Contrastive pre-training functions by contrasting instances at various scales within a graph. For each target instance, contrastive methods sample positive and negative instances, aiming to bring the positive instances closer to the target in the latent space while distancing the negative instances, as shown in Fig.~\ref{fig. pre-training}(b). Formally, for each target instance $o$ in the pre-training data $\bT_\text{pre}$, the contrastive task samples or constructs a set of positive instances $\mathcal{P}_o$, along with a set of negative instances $\mathcal{N}_o$. The contrastive loss function can be generally formulated as
\begin{align}\label{eq.contrastive-loss1}
\textstyle -\sum_{o\in \bT_\text{pre}}\ln\frac
{\sum_{a\in \mathcal{P}_o}\exp\left(\frac{\mathtt{sim}(\vec{h}_{a}, \vec{h}_{o})}{\tau}\right)}
{\sum_{a\in \mathcal{P}_o}\exp\left(\frac{\mathtt{sim}(\vec{h}_{a}, \vec{h}_{o})}{\tau}\right)+\sum_{b\in \mathcal{N}_o}\exp\left(\frac{\mathtt{sim}(\vec{h}_{b}, \vec{h}_{o})}{\tau}\right)}
\end{align}
where $\mathtt{sim}(\cdot,\cdot)$ represents a similarity function such as cosine. 

We summarize contrastive strategies for graph pre-training in Table~\ref{table.contrastive}, and discuss them based on their instance scale, augmentation techniques, and types of graph.

\stitle{Instance scale.}
First, the \emph{scale of instances} used for contrasting varies, ranging from individual nodes to broader subgraphs, or even the entire graph. One one hand, \emph{node-level} approaches provide a more granular treatment focusing on the representations of each individual node. An example is GRACE \cite{zhu2020deep}, which generates two augmented views by randomly corrupting the original graph. Then, for each target node in one view, it treats the corresponding node from the other view as positive, while others as negative. 
On the other hand, \emph{graph-level} approaches expand the scope to (sub)graphs, thereby capturing global structural patterns. 
GCC \cite{qiu2020gcc} conducts random walk in an $r$-hop ego network to obtain a target subgraph instance alongside its positive instances. Negative instances are acquired by performing random walk in an $r'$-hop ego network of the same ego node, for some $r\neq r'$. Similarly, GraphCL \cite{you2020graph} generates graph views with various data augmentations and treats graph views augmented from the same graph as positive instances, while those from different graphs as negative instances.
Extending from the graph level to the \emph{dataset level}, GraphLoG \cite{xu2021self} introduces hierarchical prototypes to capture not only the local-instance structure among correlated graphs, but also the global-semantic structure across all graphs in the dataset. 
Moreover, there also exist \emph{cross-scale} approaches contrasting between node- and graph-level instances, such as 
DGI \cite{velivckovic2018deep} and InfoGraph \cite{sun2019infograph}. They both maximize the mutual information between the local node representation and the global graph representation, but differ how they generate negative instances. On the other hand, Subg-Con \cite{jiao2020sub} captures local structure information by contrasting central nodes with their contextual subgraphs for improved scalability on very large graphs.

\stitle{Augmentation.} To sample the positive or negative instances, different augmentation techniques have been proposed for learning robust representations. The most straightforward augmentation strategy perturbs or samples the graphs in a uniformly random fashion \cite{velivckovic2018deep,zhu2020deep}, while recent studies turn to a more strategic selection of augmentations. 
MVGRL \cite{hassani2020contrastive} applies graph diffusion to transform the adjacency matrix as structure-space augmentation. 
Moreover, JOAO \cite{you2021graph} and GCGM \cite{ijcai2024gcgm} attempt to automatically select the most challenging augmentations adapted to the current loss, while
GCA \cite{zhu2021graph} adapts each graph's augmentation according to the importance of node or edge features.
On the other hand, \etal~\cite{you2022bringing} propose a learnable generator for graph augmentation. 
In contrast, SimGRACE \cite{xia2022simgrace} does not apply any augmentation to the original graph; instead, it perturbs the graph encoder to produce two correlated views. 

\stitle{Graph types.} In addition to general homogeneous graphs, contrastive methods are also widely applied across a diverse range of graph types, including heterogeneous graphs, dynamic graphs and 3D graphs \cite{batzner20223,liu2022spherical}.
For \emph{heterogeneous graphs}, HeCo \cite{wang2021self} contrasts between a schema view and a meta-path view to capture complex semantics through local and high-order structures on a heterogeneous graph. 
CPT-HG \cite{jiang2021contrastive} propose two contrastive strategies at the relation level and subgraph level. The former considers two types of negative samples from inconsistent relations and disconnected nodes, respectively, while the latter leverage meta-graphs \cite{fang2016semantic} to construct positive and negative samples.
PT-HGNN \cite{jiang2021pre} proposes a dual contrastive mechanism focusing on node and schema level tasks. Node level negative samples are pairs of nodes that match specific relations, while schema level negative samples are generated under the guidance of the network schema.
For \emph{dynamic graphs}, DDGCL \cite{tian2021self} treats a time-weighted subgraph view of a target node as postive samples, whereas taking its structurally shuffled temporal subgraph and unweighted subgraph views as negative. 
Furthermore, SelfRGNN \cite{sun2022self} proposes a Riemannian GNN with time-varying curvature, and further leverages the functional curvature over time to obtain auxiliary views without generating new graphs as augmentations.
Additionally, to learn long-short term evolution patterns and distinct structural patterns in dynamic graphs, CPDG \cite{bei2023cpdg} designs a structural-temporal sampler with temporal-aware sampling probability to extract more meaningful contextual subgraphs as positive and negative instances. 
For \emph{3D graph structures}, GearNet \cite{zhang2022protein} pre-trains protein representations using their 3D structures, 
positing that biologically relevant substructures are positioned near each other, while those that are unrelated are mapped farther apart.

\begin{table}[tbp] 
    \centering
    \small
    \addtolength{\tabcolsep}{-1mm}
    \caption{Summary of generative graph pre-training.\label{table.generative}}
    \resizebox{1\linewidth}{!}{
    \begin{tabular}{@{}c|ccccccc@{}}
    \toprule
    \multirow{3}{*}{\textbf{Method}} & \multicolumn{6}{c}{\textbf{Reconstruction objective}} & \multirow{3}{*}{\textbf{\makecell{Graph\\type}}}\\
    & \makecell{node\\feat.} & \makecell{node\\deg.} & edge & \makecell{adj.\\matrix} & \makecell{graph\\feat.} & \makecell{{other}\\{info.}}\\\midrule
    VGAE \cite{kipf2016variational}
    & $\times$
    & $\times$
    & $\times$
    & $\checkmark$
    & $\times$
    & $\times$
    & general
    \\
    GPT-GNN \cite{hu2020gpt}
    & $\checkmark$
    & $\times$
    & $\checkmark$
    & $\times$
    & $\times$
    & $\times$
    & general
    \\
    {MaskGAE \cite{li2023s} }
    & $\times$
    & $\checkmark$
    & $\checkmark$
    & $\times$
    & $\times$
    & $\times$
    & general
    \\
    {NWR-GAE \cite{tang2022graph}}
    & $\checkmark$
    & $\checkmark$
    & $\times$
    & $\times$
    & $\times$
    & $\times$
    & general
    \\
    LaGraph \cite{xie2022self}
    & $\checkmark$
    & $\times$
    & $\times$
    & $\times$
    & $\checkmark$
    & $\times$
    & general
    \\
    GraphMAE \cite{hou2022graphmae}
    & $\checkmark$
    & $\times$
    & $\times$
    & $\times$
    & $\times$
    & $\times$
    & general
    \\
    GraphMAE2 \cite{hou2023graphmae2}
    & $\checkmark$
    & $\times$
    & $\times$
    & $\times$
    & $\times$
    & $\times$
    & general
    \\
    Liu \etal~\cite{liu2022mask}
    & $\checkmark$
    & $\times$
    & $\times$
    & $\times$
    & $\times$
    & $\times$
    & KG
    \\
    Wen \etal~\cite{zhang2023structure}
    & $\checkmark$
    & $\times$
    & $\checkmark$
    & $\times$
    & $\times$
    & $\times$
    & KG
    \\
     {MPKG \cite{fan2023zero}}
    & $\checkmark$
    & $\times$
    & $\checkmark$
    & $\times$
    & $\times$
    & $\checkmark$
    & KG    
    \\    
    {PT-DGNN \cite{chen2022pre}}
    & $\times$
    & $\times$
    & $\checkmark$
    & $\times$
    & $\times$
    & $\times$
    & dynamic
    \\      
    {STEP \cite{shao2022pre}}
    & $\checkmark$
    & $\times$
    & $\times$
    & $\times$
    & $\times$
    & $\times$
    & dynamic
    \\
    {PMGT \cite{liu2021pre}}
    & $\checkmark$
    & $\times$
    & $\checkmark$
    & $\times$
    & $\times$
    & $\checkmark$
    & MMG
    \\      
    {ColdGPT \cite{cao2023multi}}
    & $\checkmark$
    & $\times$
    & $\times$
    & $\times$
    & $\times$
    & $\checkmark$
    & MMG
    \\
    \bottomrule
    \end{tabular}}
\end{table}

\subsubsection{Generative Strategies}
Concurrent to the development of contrastive strategies, generative methods present a different perspective for graph pre-training. These methods aim to reconstruct parts of the graph, unlike contrastive methods that rely on discriminating different instances on the graph. Various reconstruction objectives have been proposed, including the adjacency matrix \cite{kipf2016variational}, masked edges \cite{hu2020gpt} or node features \cite{hou2022graphmae}. These objectives involve structure reconstruction or feature reconstruction, with many studies employing objectives from both categories, as summarized in Table~\ref{table.generative}. Additionally, similar to contrastive methods, diverse generative methods have also been proposed for various types of graph. 

\stitle{Structure reconstruction} focuses on the graph topology, aiming to regenerate the entire graph structure or part of it, compelling the graph encoder to preserve the inherent structural patterns. For instance, 
VGAE \cite{kipf2016variational} attempt to reconstruct the adjacency matrix using the inner product of node embeddings. However, it mainly captures one-hop connectivity patterns, overlooking complex high-order structural information. 
Alternatively, GPT-GNN \cite{hu2020gpt} choose to mask a portion of the edges, and aim to reconstruct these missing edges using the remaining unmasked graph structure. 
Similarly, MaskGAE \cite{li2023s} masks a subset of edges and aims to reconstruct both the masked edges and the degrees of the associated nodes.
Beyond reconstructing individual edges, NWR-GAE \cite{tang2022graph} reconstructs the entire neighborhood of a node, including node degrees and the distribution of neighbors' features, offering a more comprehensive insight into the graph structure. 

\stitle{Feature reconstruction} shifts the focus to generating node features in the original or latent space. 
LaGraph \cite{xie2022self} introduces a predictive framework to perform unobserved latent graph prediction, employing node masking to compute the invariance term in its predictive objective.
Meanwhile, GraphMAE \cite{hou2022graphmae} turns to masked node feature reconstruction, proposing a masking strategy and error metric designed for robust training.
GraphMAE2 \cite{hou2023graphmae2} further introduces two decoding strategies, namely, multi-view random re-mask decoding and latent representation prediction, to regularize feature reconstruction and minimize overfitting. 

\stitle{Graph types.}
Different generative pre-training strategies have been developed for diverse types of graph. For \emph{knowledge graphs} (KGs), Liu \etal~\cite{liu2022mask} tackle complex logical queries by framing it as a masked entity prediction problem. It randomly samples subgraphs from a knowledge graph and masks certain entities for prediction. Besides masking entities, Wen \etal~\cite{zhang2023structure} mask some relations to derive a more comprehensive understanding of semantic knowledge. Moreover, MPKG \cite{fan2023zero} integrates four pretext tasks: knowledge reconstruction, higher-order neighbor reconstruction, feature reconstruction, and relation adaptation to capture heterogeneous item semantics, particularly enhancing performance in zero-shot scenarios. 
For \emph{dynamic graphs}, PT-DGNN \cite{chen2022pre} first samples subgraphs in a time-aware manner as the input to pre-training, and then predicts unobserved edges using observed one to capture the evolutionary traits of the network. On the other hand, STEP \cite{shao2022pre}  masks node features over a time period for prediction, providing an alternative approach to understanding graph evolution. 
For \emph{multi-modal graphs} (MMGs), PMGT \cite{liu2021pre} employs graph structure reconstruction and masked node feature reconstruction, integrating item relations and multi-modal side information. Similarly, ColdGPT \cite{cao2023multi} designs multiple pretext tasks from various data sources, such as item content, historical purchase sequences, and review texts.

\subsubsection{Integration with Pre-trained Large Models}
Large language models (LLMs) have recently emerged as a powerful and widely adopted paradigm across the broader fields of machine learning and artificial intelligence due to their emergent capabilities. Inspired by their remarkable success, initial efforts have been made to incorporate LLMs into graph pre-training. These explorations predominantly focus on text-attributed graphs (TAGs), where each node or edge is accompanied by textual descriptions that can be naturally processed by LLMs.
To integrate both structural and semantic information within TAGs, contrastive pre-training methods have been adopted in several studies \cite{wen2023augmenting, wen2023prompt, tang2023graphgpt, brannon2023congrat}. Similar to Contrastive Language-Image Pretraining (CLIP) \cite{radford2021learning}, their core principle is to align text embeddings, derived from a language model, with their corresponding node embeddings produced by a graph encoder. 
Alternatively, Patton \cite{jin2023patton} combines masked language modeling \cite{kenton2019bert} and masked node prediction simultaneously to pre-train the LLM and the graph encoder. LLMs can also be used at various stages of graph pre-training. For instance, both GALM \cite{xie2023graph} and InstructGLM \cite{ye2023natural} employ link prediction as a pretext task, but their application of LLMs differ. Specifically, GALM feeds LLM-encoded node embeddings into the graph encoder to incorporate graph structural information, using link prediction to pre-train both the LLM and the graph encoder. In contrast, InstructGLM relies on natural language to  describe the graph structures and employs the LLM to encode node embeddings.
Moreover, instead of relying on unsupervised pre-training strategies, SimTeG \cite{duan2023simteg} first fine-tunes a pre-trained language model with supervision from a downstream task, and then utilizes the fine-tuned model to obtain node embeddings, which can be subsequently used by any graph encoder for training on the same task.



\subsection{Downstream Adaptation by Fine-tuning}\label{sec:pre-train:fine-tune}
The pre-trained graph encoder entails prior knowledge about the intrinsic properties of the graphs used in the pre-training stage. This prior can be transferred to various downstream tasks by initializing a downstream model with the pre-trained weights. 
To tailor the initial model to each specific task, it is equipped with a task-specific projection head, and undergoes an adaptation stage known as \emph{fine-tuning}. 
During fine-tuning, for each task, the model is further trained with task-specific annotations, which can update the parameters in both the pre-trained model and the task head in the process  \cite{qiu2020gcc,hu2020gpt}, as shown in Fig.~\ref{fig. pre-training}(c), left. 

However, adaptation by fine-tuning, while intuitive and straightforward, has its limitations \cite{liu2023pre}. First, an objective gap often exists between the pretext tasks in pre-training and the actual downstream tasks, which can compromise the effectiveness of the adaptation process. 
Second, updating all parameters can be inefficient, especially in the case of large pre-trained models.
In our context of few-shot learning, fine-tuning may lead to overfitting due to the limited number of task-specific labels \cite{liu2023graphprompt}.

\subsection{Parameter-efficient Downstream Adaptation}\label{sec:pre-train:param-efficient}
To mitigate the problems of adaptation by fine-tuning, recent efforts have focused on parameter-efficient adaptation, which avoids updating all parameters in the pre-trained graph encoder. Key strategies include prompt tuning and parameter-efficient fine-tuning.

\begin{table*}[tbp]
    \centering
    \small
    \addtolength{\tabcolsep}{-1mm}%
        \caption{Summary of prompt tuning on graphs. For ease of comparison, we adapt the notations used in the original references with a consistent set of symbols. $v,u$: Nodes $v,u$; $\Vec{t}_y$: Prototype token $\vec{t}$ of class $y$;  $\Vec{h}_v$: Node $v$'s embedding; $\vec{x}_v$: Node $v$'s attribute;  $\vec{s}_v$: Subgraph of $v$'s embedding;  CL: Contrastive learning; GL: Generative learning; CLS: Classification.
        \label{table.prompt-tuning}
        }
    \resizebox{1\linewidth}{!}{
        \begin{tabular}{@{}c|c|c|c|c|c|ccc@{}}
        \toprule
        \multirow{2}*{\textbf{Paper}} & \multirow{2}*{\textbf{Template}} & {\textbf{Feature}} & {\textbf{Structure}}  & {\textbf{Multiple}}  & \textbf{Prompt} & \multicolumn{3}{c}{\textbf{Downstream Task}} \\
         & & \textbf{prompt} & \textbf{prompt} & \textbf{pretext tasks} & \textbf{Initialization} & Node & Edge & Graph \\ \midrule
        
        GPPT \cite{sun2022gppt} 
        & subgraph-token similarity: $\text{sim}(\vec{s}_v, \Vec{t}_y)$
        & input
        & $\times$
        & $\times$
        & random
        & $\checkmark$
        & $\times$
        & $\times$  
        \\ \midrule
        VPGNN \cite{wen2023voucher} 
        & node-token matching: $\text{match}(\vec{h}_v,\vec{t}_y)$
        & $\times$
        & $\checkmark$
        & $\times$
        & random
        & $\checkmark$
        & $\times$
        & $\times$
        \\\midrule
        GraphPrompt \cite{liu2023graphprompt}
        & \multirow{4}{*}{subgraph similarity: $\text{sim}(\vec{s}_u,\vec{s}_v)$}
        & readout
        & $\times$
        & $\times$
        & random
        & $\checkmark$
        & $\checkmark$
        & $\checkmark$ 
        \\
        {MOP \cite{hao2024motif}}
        & 
        & readout
        & $\times$
        & $\times$
        & random
        & $\times$
        & $\checkmark$
        & $\times$
        \\
        \makecell{
        GraphPrompt+ \cite{yu2023generalized}
        }
        & 
        & all layers
        & $\times$
        & $\times$
        & random
        & $\checkmark$
        & $\checkmark$
        & $\checkmark$ 
        \\
        {ProNoG \cite{yu2024non}}
        & 
        & readout
        & $\times$
        & $\times$
        & conditional 
        & $\checkmark$
        & $\checkmark$
        & $\checkmark$ 
        \\
        \midrule
        {MDGPT \cite{yu2024text}}
        & \multirow{3}{*}{
        node similarity: $\text{sim}(\vec{h}_u,\vec{h}_v)$
        }
        & readout
        & $\times$
        & $\times$
        & pretext tokens
        & $\checkmark$
        & $\checkmark$
        & $\checkmark$ 
        \\
        MultiGPrompt \cite{yu2023multigprompt}
        & 
        & all layers
        & $\times$
        & $\checkmark$
        & pretext tokens
        & $\checkmark$
        & $\checkmark$
        & $\checkmark$ 
        \\
        {HetGPT \cite{ma2024hetgpt}}
        & 
        & input
        & $\times$
        & $\times$
        & random
        & $\checkmark$
        & $\times$
        & $\times$
        \\
        \midrule
        GPF \cite{fang2022universal}  
        & \multirow{2}{*}{universal feature/spectral space} 
        & input
        & $\times$
        & $\times$
        & random
        & $\checkmark$
        & $\checkmark$
        & $\checkmark$ 
        \\
        {IGAP \cite{yan2024inductive}}
        & 
        & signal
        & $\times$
        & $\times$
        & random
        & $\checkmark$
        & $\times$
        & $\checkmark$
        \\\midrule
        SGL \cite{zhu2023sgl} 
        & dual-template: $\text{CL}(\vec{h}_u,\vec{h}_v), \text{GL}(\vec{x}_v, \vec{\tilde{x}}_v)$
        & $\times$
        & $\checkmark$
        & $\checkmark$
        & random
        & $\times$
        & $\times$
        & $\checkmark$
        \\ \midrule
        HGPrompt \cite{yu2023hgprompt} 
        & dual-template: $\text{sim}(\vec{s}_u,\vec{s}_v)$, graph template 
        & readout
        & $\times$
        & $\times$
        & random
        & $\checkmark$
        & $\checkmark$
        & $\checkmark$ 
        \\
        \midrule
        SAP \cite{ge2023enhancing} 
        & view similarity: $\text{sim}(\text{MLP}(X), \text{GNN}(X,A))$ 
        & $\times$
        & $\checkmark$
        & $\checkmark$
        & random
        & $\checkmark$
        & $\times$
        & $\checkmark$
        \\\midrule
        ULTRA-DP \cite{chen2023ultra} 
        &  node-node{/}group similarity:  
            $\text{sim}(\vec{h}_u,\vec{h}_v)$ 
        & input
        & $\checkmark$        
        & $\checkmark$
        & random
        & $\checkmark$
        & $\checkmark$
        & $\times$
        \\\midrule
        VNT \cite{tan2023virtual}  
        & \makecell{ 
            node attribute reconstruction: $\text{MSE}(\vec{x}_v, \vec{\tilde{x}}_v)$ \\
            structure recovery: $\text{MSE}(\{\vec{h}_u,\vec{h}_v\})$
            }
        & input
        & $\times$
        & $\times$
        & meta-trained
        & $\checkmark$
        & $\times$
        & $\times$ 
        \\\midrule
        ProG \cite{sun2023all}  
        & subgraph classification: $\text{CLS}(\vec{s})$
        & $\times$
        & $\checkmark$
        & $\times$
        & meta-trained
        & $\checkmark$
        & $\checkmark$
        & $\checkmark$ 
        \\\midrule
        {DyGPrompt \cite{yu2024dygprompt} }
        & \multirow{2}{*}{temporal node similarity: $\text{sim}(\vec{h}_{t,u},\vec{h}_{t,v})$}
        & input
        & $\times$
        & $\times$
        & conditional 
        & $\checkmark$
        & $\checkmark$
        & $\times$ 
        \\
        {TIGPrompt \cite{chen2024prompt}}
        & 
        & input
        & $\times$
        & $\times$
        & \makecell{time-based}
        & $\checkmark$
        & $\checkmark$
        & $\times$ 
        \\
        \bottomrule
        \end{tabular} 
    }
\end{table*}

\begin{table}[tbp]
    \centering
    \small
    \addtolength{\tabcolsep}{-1mm}%
        \caption{Summary of prompt tuning on text-attributed graphs.
        \label{table.prompt-tuning-tag}
        }
    \resizebox{1\linewidth}{!}{
        \begin{tabular}{@{}c|cc|c|ccc@{}}
        \toprule
        \multirow{2}*{\textbf{Paper}} & \multicolumn{2}{c|}{\textbf{Instruction}} & \textbf{Learnable}  & \multicolumn{3}{c}{\textbf{Downstream Task}} \\
         & Text & Graph & \textbf{prompt} & Node & Edge & Graph \\ \midrule
        {G2P2 \cite{wen2023augmenting}}
        & $\checkmark$
        & $\times$
        & vector
        & $\checkmark$
        & $\times$
        & $\times$
        \\
        {G2P2* \cite{wen2023prompt}}
        & $\checkmark$
        & $\times$
        & condition-net
        & $\checkmark$
        & $\times$
        & $\times$
        \\
        {GraphGPT \cite{tang2023graphgpt}}
        & $\checkmark$
        & $\checkmark$
        & $\times$
        & $\checkmark$
        & $\times$
        & $\times$
        \\
        {InstructGLM \cite{ye2023natural}}
        & $\checkmark$
        & $\checkmark$
        & $\times$
        & $\checkmark$
        & $\times$
        & $\times$
        \\
        {GIMLET \cite{zhao2023gimlet}}
        & $\checkmark$
        & $\checkmark$
        & $\times$
        & $\times$
        & $\times$
        & $\checkmark$
        \\
        {OFA \cite{liu2023one}}   
        & $\checkmark$
        & $\checkmark$
        & $\times$
        & $\checkmark$
        & $\checkmark$
        & $\checkmark$ 
        \\
        {HiGPT \cite{tang2024higpt}}   
        & $\checkmark$
        & $\checkmark$
        & $\times$
        & $\checkmark$
        & $\times$
        & $\times$
        \\
        \bottomrule
        \end{tabular} 
    }
\end{table}

\subsubsection{Prompt Tuning on Graphs}
Originating in the field of natural language processing (NLP), \emph{prompt-based learning} has demonstrated its efficacy in adapting pre-trained language models to diverse language tasks \cite{liu2023pre}. In essence, a prompt vector is introduced to modify or reformulate the original input for the pre-trained model, drawing the downstream task closer to the pretext tasks under a unified template. Notably, the prompt vector has a negligible size relative to the pre-trained parameters. During the adaptation process, only a very small prompt vector is tuned without updating the pre-trained model, making the approach parameter efficient.  
Recently, prompt tuning have been extended to graph learning \cite{sun2022gppt,liu2023graphprompt}. Building on a unified template that aligns the pretext and downstream losses, these methods design prompts to further narrow the gap between pretext and downstream tasks, as illustrated in Fig.~\ref{fig. pre-training}(c), right. 
Specific approaches vary in the choice of the template and prompt design, as summarized in Tables~\ref{table.prompt-tuning}\&~\ref{table.prompt-tuning-tag}.

\stitle{Prompt template.} Similarity calculation has been widely adopted \cite{sun2022gppt,liu2023graphprompt,yu2023generalized,hao2024motif,yu2023multigprompt,chen2023ultra,yu2023hgprompt,yu2024dygprompt,yu2024text,yu2024non} to unify three major tasks: node classification (NC), link prediction (LP), and graph classification (GC). For instance, GraphPrompt \cite{liu2023graphprompt} reformulates LP as calculating the similarity between the ego subgraphs of two nodes. Then, it applies the same template to NC and GC, treating them as calculating the similarity between the subgraph of the target instance (using an ego network for NC, or the target graph itself for GC) and the prototypical subgraph for each class. While GraphPrompt only adopts LP as the pretext task, GraphPrompt+ \cite{yu2023generalized} illustrates that other prevailing contrastive pretext tasks can also be redefined using subgraph similarity. Likewise, SGL \cite{zhu2023sgl}, SAP \cite{ge2023enhancing} and VPGNN \cite{wen2023voucher} also employ a form of similarity or matching score to unify pretext and downstream tasks. In particular, SGL further incorporates a generative pretext task based on mask node feature reconstruction, which is employed as an additional template for downstream prompt learning. 
In contrast, GPF \cite{fang2022universal} and IGAP \cite{yan2024inductive} do not employ an explicit template to align task formats. Instead, GPF assumes a universal feature space as an implicit template to unify different tasks. Meanwhile,  IGAP posits a universal graph signal and spectral space as an implicit template.

\stitle{Prompt design} can be further categorized into \emph{feature prompt} and 
\emph{structure prompt}.
On one hand, similar to prompts in NLP \cite{liu2023pre}, feature prompts modify node features, reformulating the original downstream tasks to narrow the gap from the pre-trained model. Some studies incorporate feature prompts in the input layer \cite{sun2022gppt,fang2022universal,chen2023ultra} to explicit modify the input in a task-specific manner, while others apply them to the readout layer to guide subgraph readout \cite{liu2023graphprompt,hao2024motif,yu2023hgprompt,yu2024non} or all layers of the pre-trained model \cite{yu2023generalized,yu2023multigprompt}.  Moreover, from a graph spectral perspective, feature prompts can be applied to graph signals to reduce the spectral  gap between various graphs \cite{yan2024inductive}.
On the other hand, structure prompts act as a virtual graph structure to reformulate the implicit structural input for downstream tasks. Structure prompts can take the form of virtual nodes or graphs representing class prototypes for classification tasks \cite{zhu2023sgl,ge2023enhancing,chen2023ultra,wen2023voucher}, or virtual nodes that augment link prediction \cite{chen2023ultra}. ProNoG \cite{yu2024non} generates a series of prompts for each node using a condition-net, aiming to adapt to node-specific structural patterns on non-homophilic graphs.
Furthermore, some studies propose utilizing multiple pretext tasks \cite{zhu2023sgl,ge2023enhancing,chen2023ultra} to acquire comprehensive prior knowledge from diverse perspectives. However, directly combining various pretext tasks may lead to task interference \cite{yu2023multigprompt}, compromising the complementary benefits of employing multiple tasks. To mitigate this issue, MultiGPrompt \cite{yu2023multigprompt} introduces a composed prompt, which is composed of a series of pretext tokens used during pre-training to promote cooperation among the multiple pretext tasks.
Additionally, to leverage prior knowledge across graphs from multiple domains, MDGPT \cite{yu2024text} employs a mixing prompt that aggregates a set of domain tokens to bridge the semantic gap across different domains. 

\stitle{Graph types.} Prompt-based learning has also been considered on various types of graph.  
For \emph{heterogeneous graphs}, HGPrompt \cite{yu2023hgprompt} extends GraphPrompt, utilizing an additional graph template to convert a heterogeneous graph into homogeneous ones, bridging the gap between the two kinds of graph. Additionally, HetGPT \cite{ma2024hetgpt} modifies each type of nodes with a type-specific prompt.
For \emph{dynamic graphs}, TIGPrompt \cite{chen2024prompt} proposes node-specific prompts based on timestamps, while 
DyGPrompt \cite{yu2024dygprompt} 
generates conditional prompts for each node and timestamp via dual condition-nets to capture the mutual influence between node and time features. 
For \emph{text-attributed graphs}, pre-training often utilizes pre-trained language models. Consequently, in the downstream adaptation stage, various  prompt-based learning methods have been specifically designed for these language models, as summarized in Table~\ref{table.prompt-tuning-tag}.
For instance, G2P2 \cite{wen2023augmenting} appends discrete language prompts to textual class labels in zero-shot node classification and tunes learnable prompt vectors in few-shot settings. Moreover, inspired by CoCoOp \cite{zhou2022conditional}, G2P2* \cite{wen2023prompt} utilizes node embeddings as conditions to train a learnable conditional prompt for the language model.
Other works \cite{tang2023graphgpt,ye2023natural,zhao2023gimlet,tang2024higpt} generally describe graph structures using  textual instructions for LLMs.
Conversely, 
OFA \cite{liu2023one} introduces prompt nodes to unify various downstream tasks by fusing the node-of-interest subgraph, prompt node, and class nodes into a single graph.

\subsubsection{Parameter-Efficient Fine-Tuning (PEFT)}
These methods only tune a selected subset of parameters from the original pre-trained model or from newly added modules. Hence, the number of parameters requiring updating is significantly reduced, which proves more efficient and feasible for few-shot learning.
Prominent PEFT techniques include \emph{adapter tuning} \cite{houlsby2019parameter} and \emph{Low-Rank Adaptation} (LoRA) \cite{hu2021lora}.
Adapter tuning \cite{houlsby2019parameter} inserts small neural network modules called adapters into certain layers of a pre-trained model, while LoRA \cite{hu2021lora} utilizes low-rank matrices to approximate parameter updates. In both techniques, only the parameters of the new modules are updated during fine-tuning, while the original pre-trained weights remain frozen.

For pre-trained graph models, AdapterGNN \cite{li2023adaptergnn} introduces adapter-based fine-tuning into graph neural networks. This is achieved  via two parallel adapters that process inputs before and after message passing. In a similar spirit, G-Adapter \cite{gui2023g} refines the graph transformer architecture with adapters, and further incorporates structural information into the adapter. In a different graph-assisted application, MolCa \cite{liu2023molca} focus on the molecule-to-text generation task and propose molecular graph language modeling. They design a cross-modal projector aimed at bridging the gap between graph structural and textual representations, while integrating a uni-modal LoRA module into their language model.

\subsection{Discussion}
Unlike meta-learning approaches that require a large volume of labeled data to construct meta-learning tasks, pre-training approaches employ self-supervised pretext tasks on unlabeled data. 
Therefore, pre-training approaches are more effective in scenarios where labeled data are limited to novel tasks without a pre-existing set of annotated tasks. 
In contrast, when such an annotated base set is available, meta-learning tends to perform better as it can leverage related meta-training tasks derived from the base set. However, the dependency on related tasks suggests that meta-learning approaches cannot be easily extended to a wide array of downstream tasks, rendering them less general than pre-training approaches.

Adapting a pre-trained model to downstream tasks by fine-tuning is an intuitive strategy, but it can be ineffective due to the objective gap and overfitting to few-shot labels, or costly for a large model. Parameter-efficient adaptation strategies, including prompt tuning, adapter tuning and LoRA, present a more promising direction for few-shot learning on graphs.

\section{Hybrid Approaches}\label{sec.hybrid}

Meta-learning and pre-training embody two distinct paradigms to learn prior knowledge, each with its unique advantages and disadvantages, as previously discussed.  
In situations where both a vast set of unlabeled data for pre-training and an extensively annotated base set for meta-learning are available, hybrid approaches that embrace both paradigms may be a natural and strategic choice for few-shot learning.
Hence, we can leverage the respective strengths of each, not only exploiting the unlabeled data to learn a general task-agnostic prior via pre-training, but also drawing insights from similar meta-training tasks. 

\begin{figure}[t]
\centering
\includegraphics[width=0.85\linewidth]{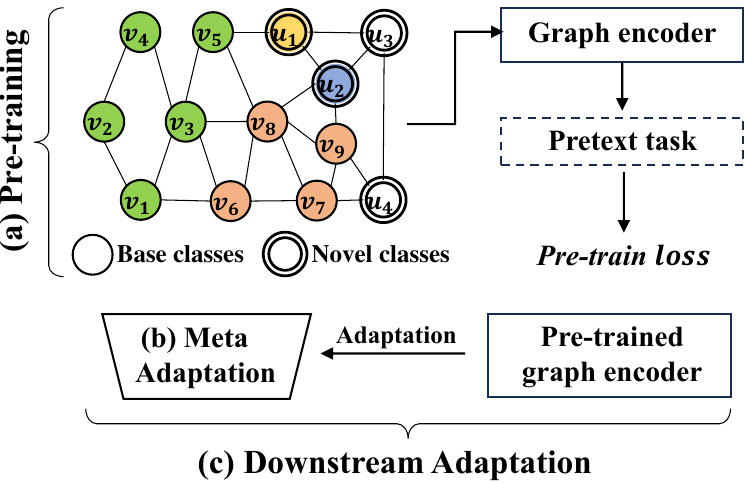}
\caption{An illustration of hybrid approaches that integrate meta-learning and pre-training for few-shot learning on graphs.}
\label{fig.hybrid-framework}
\end{figure}

As depicted in Fig.~\ref{fig.hybrid-framework}(a), hybrid approaches adopt typical contrastive or generative pretext tasks to pre-train a graph encoder. Subsequently, the pre-trained model is adapted in conjunction with meta-learning, as illustrated in Fig.~\ref{fig.hybrid-framework}(b,c).
Specifically, VNT \cite{tan2023virtual} employs a pre-trained model to reduce reliance on the base classes. It further proposes structure prompts, serving as virtual nodes to tailor the pre-trained node embeddings for meta-training tasks, further leveraging the insights from the base classes. 
However, VNT is specifically designed for node-level tasks only, falling short of handling  edge- or graph-level tasks.
In contrast, ProG \cite{sun2023all} reformulate the downstream node- and edge-level tasks as graph-level tasks, and proposes prompt graphs with specific nodes and structures to guide different tasks. Similar to VNT, it incorporates meta-learning to acquire initial prompts for handling different tasks.
On text-attributed graphs, G2P2* \cite{wen2023prompt} extends G2P2 \cite{wen2023augmenting} with conditional prompt tuning, which employs a meta-net to generate a distinct prompt vector for each node. The meta-net is trained using labeled data from base classes, facilitating knowledge transfer to novel classes.

\section{Future Directions}\label{sec.future}

This section aims to identify and discuss potential opportunities for few-shot learning on graphs, leveraging the insights gained from our review of problems and techniques. 

\subsection{Future Avenues in Problem Settings}\label{future.prob}

\stitle{Structure scarcity learning on graphs.}
As elaborated in Sect.~\ref{sec.structure-scarcity}, the issue of structure scarcity in graphs is prevalent across various real-world scenarios. Existing solutions \cite{liu2020towards,pan2019warm} to this challenge typically rely on meta-learning approaches, which learns a prior from tasks involving structurally rich instances. 
However, meta-learning often assumes that base and novel tasks are independent and identically distributed (i.i.d.). When dealing with structure scarcity, the base tasks are constructed from structure-rich instances, while the novels tasks are associated with structure-scarce instances. The discrepancy may lead to distinct distributions between them, violating the i.i.d.~assumption. 
One potential solution is to leverage pre-training to learn priors from the base tasks. Since downstream tasks typically differ from those used in pre-training, specialized adaptation techniques are often introduced to bridge the gap between them, rather than making the i.i.d.~assumption.

\stitle{Few-shot learning on large-scale graphs.}
Many previous works \cite{you2020graph,liu2023graphprompt} are applied to small-scale graph structures such as molecular graphs \cite{hu2020open}, which consist of tens or hundreds of nodes only. However, many real-world applications deal with much larger graph structures, such as social networks, e-commerce systems, and protein or gene networks.  Though some existing works \cite{rong2020self,duan2022comprehensive} reduce the computational cost to a certain extent, it remains a significant issue on very large graphs. Additionally, challenges persist in developing  finer-grained adaptation strategies to deal with potential variations among distant localities on a large graph. 

\stitle{Few-shot learning on complex graphs.}
Most of the current works focus on simple homogeneous graphs, commonly exemplified by 2D graphs, uni-modal graphs, and static graphs. However, the practical applicability of few-shot learning extends beyond these limitations to include more complex structures, such as 3D graphs \cite{liu2021pre3d}, multi-modal graphs \cite{li2023graphadapter}, and dynamic graphs \cite{yang2022few}, each with important applications in scientific, biomedical or social domains. This expansion in scope highlights the need for novel methodologies and techniques to effectively address the challenges posed by these complex graph structures. However, existing methods typically focus on a single aspect of these complexities or are tailored to specific applications \cite{liu2021pre3d,zhu2022unified}, lacking a unified approach that integrates various aspects and general applications. 

\stitle{Few-shot learning on cross-domain graphs.}\label{cross-domain}
Graphs are ubiquitous across various domains, from social networks and commerce networks to molecular graphs and protein-protein interaction networks. Each domain's graphs exhibit unique semantic meanings and topological structures. For example, in a social network, a node's features may represent a user's profile, while the links denote the connections between users. In contrast, in a molecular graph, a node's features represent the attributes of an atom, and the edges indicate chemical bonds. This diversity highlights the need for cross-domain few-shot learning methods. Some current solutions rely on textual descriptions on graphs to bridge different domains via LLMs \cite{liu2023one}, but this creates a heavy dependency on LLMs and does not work for text-free graphs \cite{yu2024text}. Other approaches ignore the topological differences across various domains \cite{zhao2024all,yu2024text}, calling for specialized domain adaptation techniques to mitigate both feature and structure variations across domains.

\subsection{Future Avenues in Techniques}\label{future.tech}
\stitle{Improving interpretability.}
Enhancing interpretability is a critical future direction for few-shot learning on graphs. The black-box nature of some techniques makes it challenging to understand and interpret their results.
For instance, unlike language-based prompts in NLP that can be understood by humans, most prompts designed for graphs are simply learnable vectors in the latent space. It thus becomes imperative to develop techniques that provide insights into the rationale behind predictions for few-shot tasks on graphs. Research in this direction could start with designing more interpretable prompts on graphs, explicitly explaining their relationship to node features or graph structures. 

\stitle{Foundation models on graphs.}
Foundation models pre-trained on large-scale data in NLP have shown powerful performance across various kinds of tasks in diverse application domains. Similarly, a pre-trained graph foundation model holds significant promise for diverse few-shot learning tasks on graphs. Toward graph foundation models, we aim to improve the transferability of current pre-training efforts on graphs, enabling their application in diverse domains as well as different task types. 
On one hand, improving transferability across domains requires not only mitigating the issue of domain shift on cross-domain graphs, but also bridging the gap created by heterogeneity in graph structures as discussed in Sect.~\ref{future.prob}. This heterogeneity stems from different kinds of complex structures, including 3D, multi-modal and dynamic graphs. Unifying these structures is critical to the universal applicability of graph foundation models; in particular, multi-modal graphs pave the way for integrating graph foundation models with language- and vision-based models.
On the other hand, there is a pressing need to enhance transferability across task types, extending beyond the conventional tasks of link prediction, node classification, and graph classification. More variety of task types, such as regression, graph editing and generation \cite{liu2023multi}, can be crucial to material engineering and drug development.

\section{Conclusions}\label{sec.conclusion}
In this survey, we offer a systematic review of the literature on few-shot learning on graphs. 
Specifically, we develop two comprehensive taxonomies for few-shot learning on graphs, classifying existing studies according to their problems and techniques.
Based on the problems, we divide these studies into two categories: label-scarcity problem and structure-scarcity problem. For each category, we summarize their problem definitions and their respective applications.
Based on the techniques, we categorize the literature into meta-learning, pre-training, and hybrid approaches. For each category, we synthesize representative works and discuss their advantages and disadvantages. Finally, we outline promising future directions, aiming to stimulate further exploration in this rapidly evolving field.


\bibliographystyle{IEEEtran}
\bibliography{references}

\begin{IEEEbiography}
[{\includegraphics[width=1in,height=1.25in,clip,keepaspectratio]{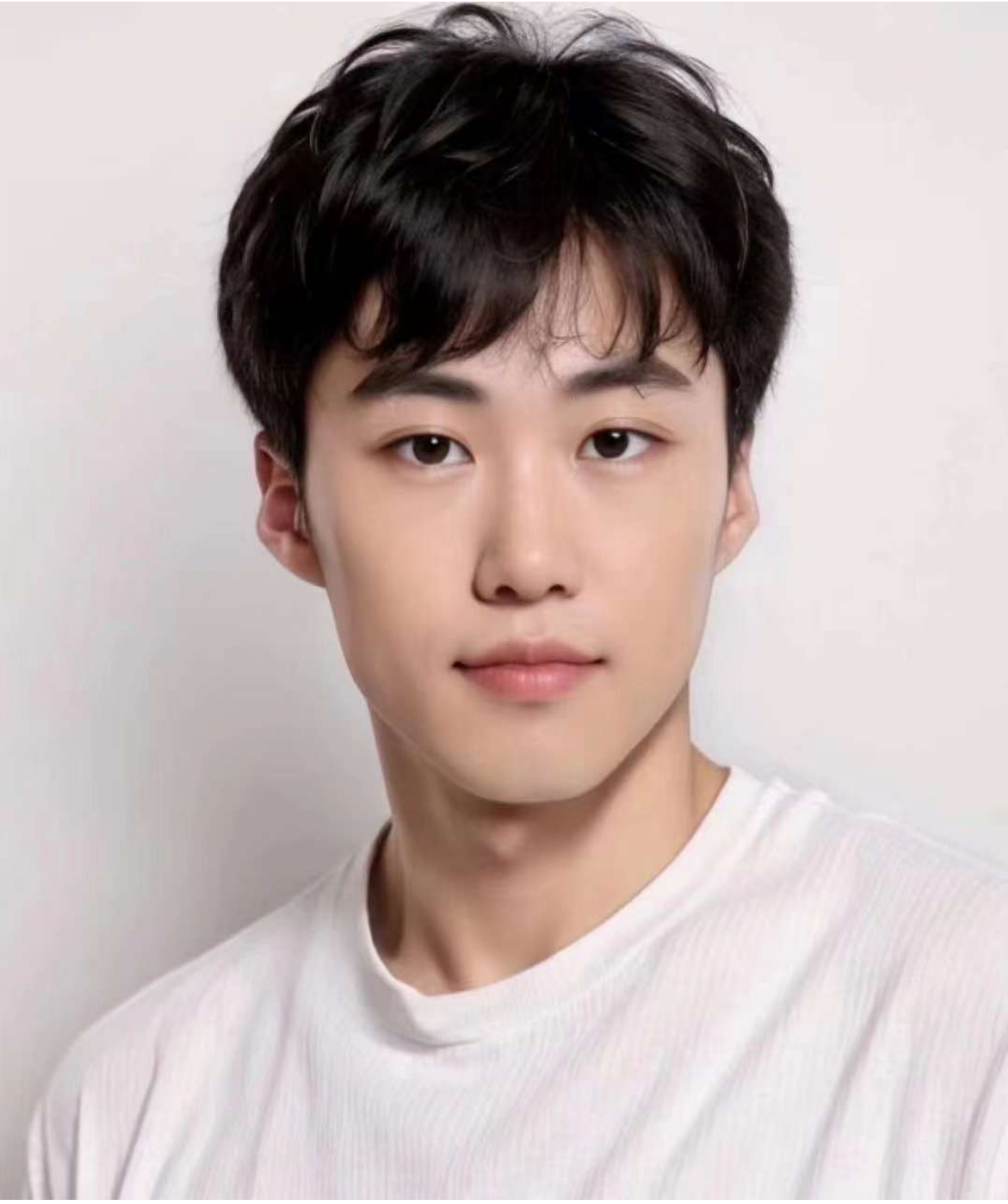}}]
{Xingtong Yu}
is currently a research scientist with the School of Computing and Information System, Singapore Management University. He received his bachelor’s degree from School of the Gifted Young, University of Science and Technology of China in 2019, and the PhD degree from the University of Science and Technology of China in 2024.  His current research focuses on graph-based machine learning, prompting on graphs, and graph foundation model. 
\end{IEEEbiography}

\begin{IEEEbiography}
[{\includegraphics[width=1in,height=1.25in,clip,keepaspectratio]{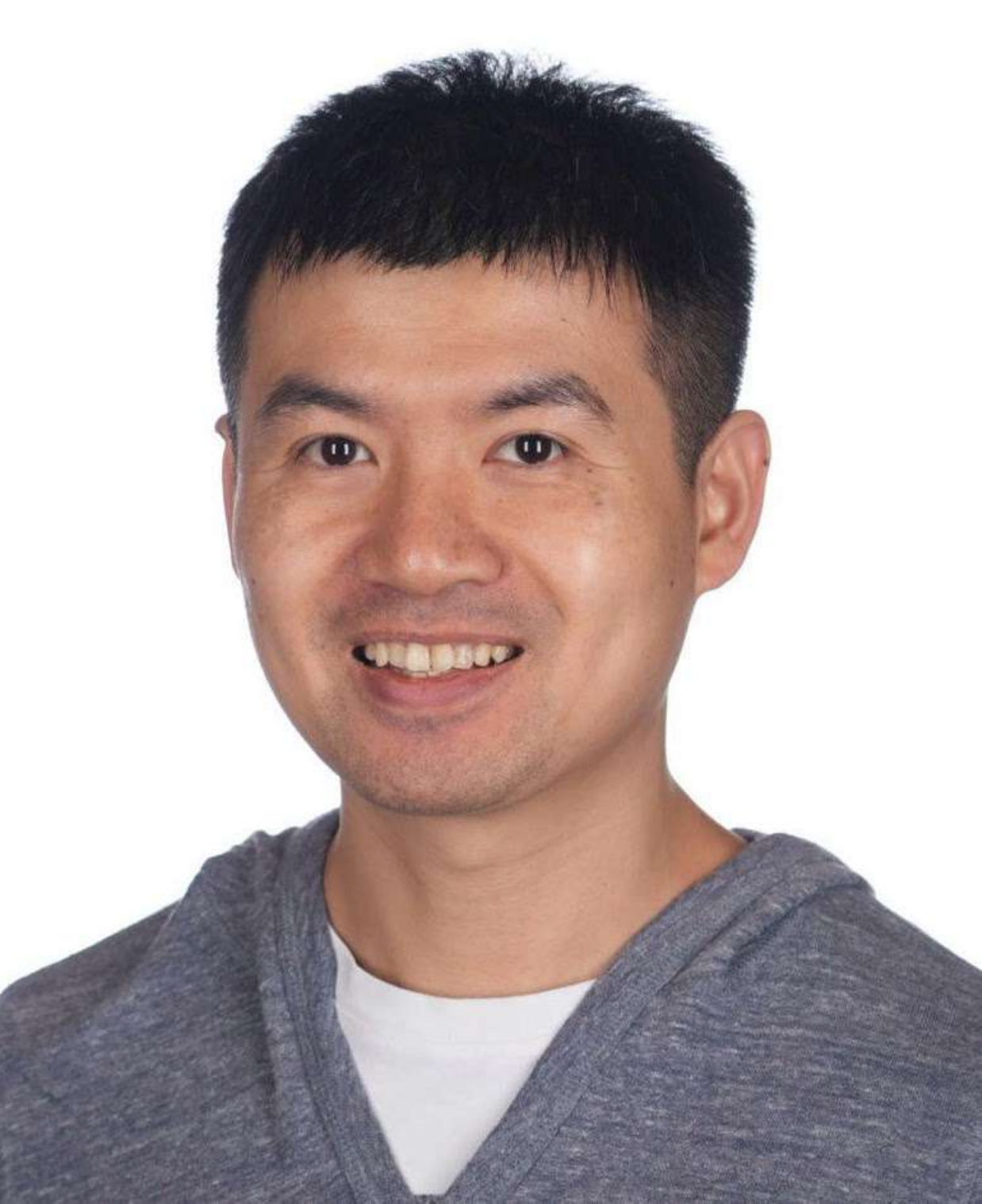}}]
{Yuan Fang}
(Senior Member, IEEE) received the bachelor’s degree in computer science from the National University of Singapore in 2009 and the PhD degree in computer science from the University of Illinois at Urbana-Champaign in 2014. He is currently an assistant professor with the School of Computing and Information Systems, Singapore Management University. His current research focuses on graph-based data mining and machine learning, and their applications.
\end{IEEEbiography}

\begin{IEEEbiography}
[{\includegraphics[width=1in,height=1.25in,clip,keepaspectratio]{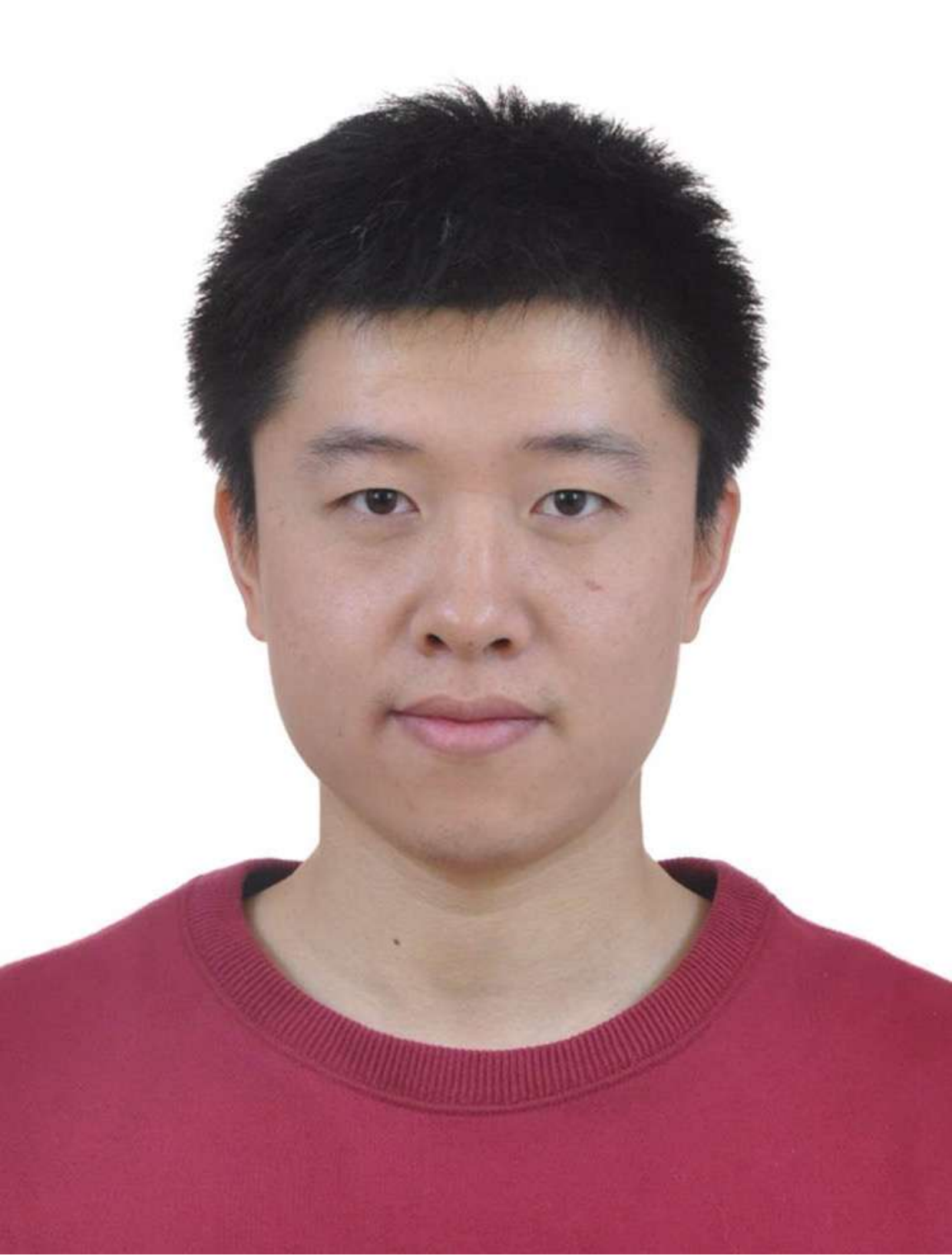}}]
{Zemin Liu}
is currently an assistant professor with the College of Computer Science and Technology, Zhejiang University. He received his Ph.D. degree in Computer Science from Zhejiang University, Hangzhou, China in 2018, and B.S. Degree in Software Engineering from Shandong University, Jinan, China in 2012. His research interests lie in graph embedding, graph neural networks, and learning on heterogeneous information networks.
\end{IEEEbiography}

\begin{IEEEbiography}
[{\includegraphics[width=1in,height=1.25in,clip,keepaspectratio]{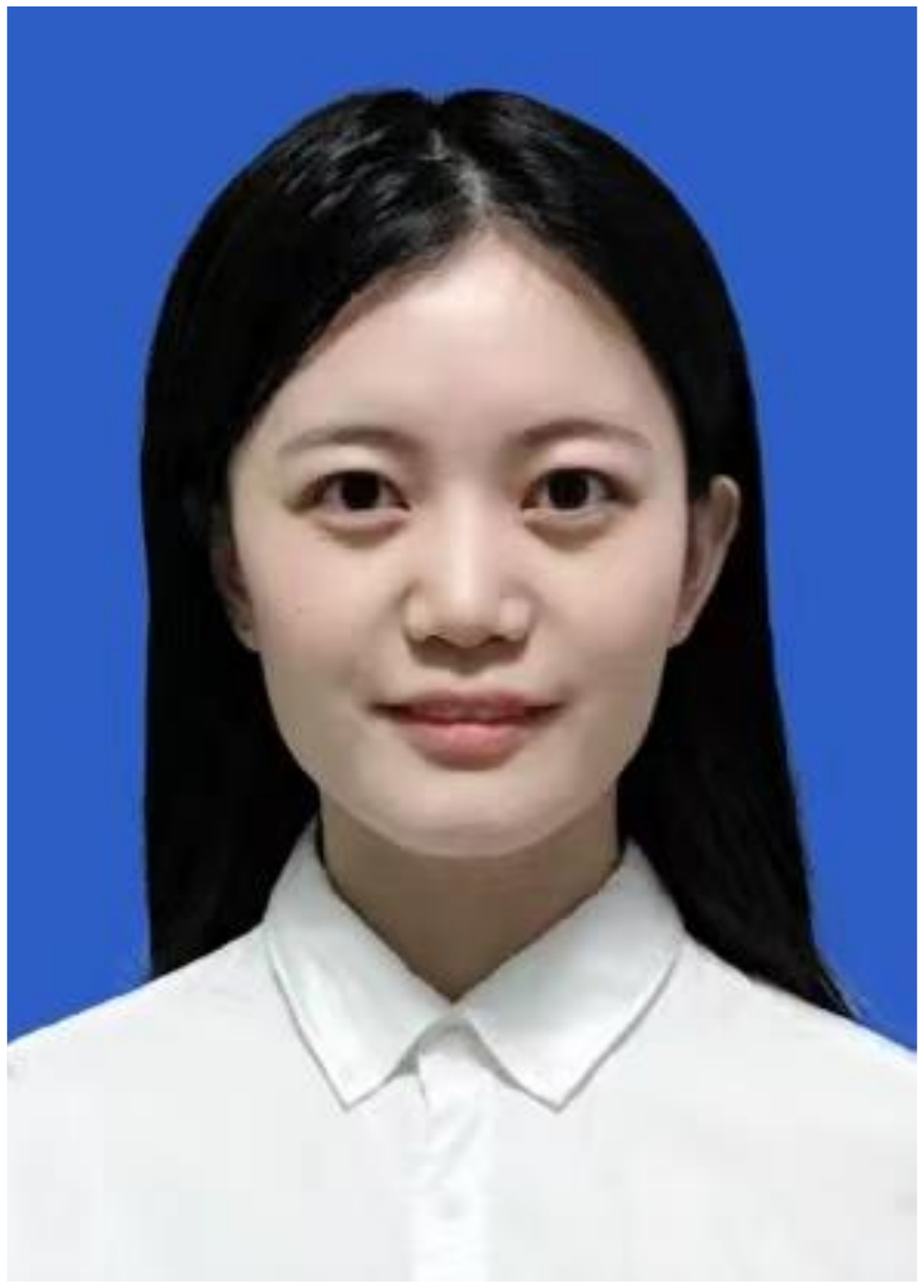}}]
{Yuxia Wu} is a research scientist at Singapore Management University. She received the Ph.D. degree from Xi'an Jiaotong University in 2023. Her research interests include natural language processing, social multimedia mining, graph mining and recommender systems. She has served as the reviewer and program committee member for multiple conferences and journals, including TPAMI, TKDE, ACL, EMNLP, ACM MM etc. 
\end{IEEEbiography}

\begin{IEEEbiography}
[{\includegraphics[width=1in,height=1.25in,clip,keepaspectratio]{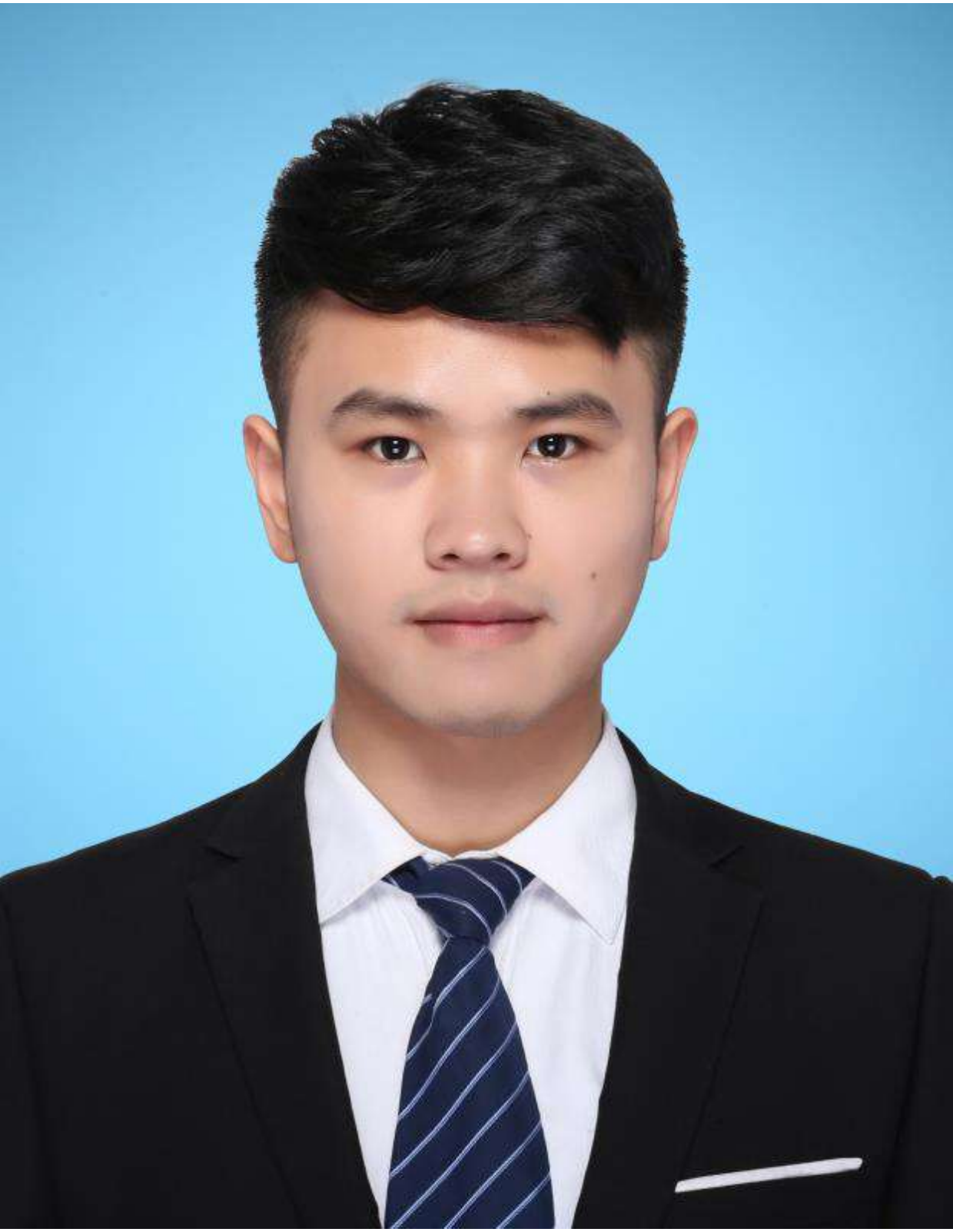}}]
{Zhihao Wen} received his Ph.D. degree in Computer Science from Singapore Management University in 2023. He is now a research fellow in the School of Computing and Information Systems, Singapore Management University. His research interests include parameter efficient fine-tuning, graph-based machine learning, data mining, as well as their applications for the Web and social media.
\end{IEEEbiography}

\begin{IEEEbiography}
[{\includegraphics[width=1in,height=1.25in,clip,keepaspectratio]{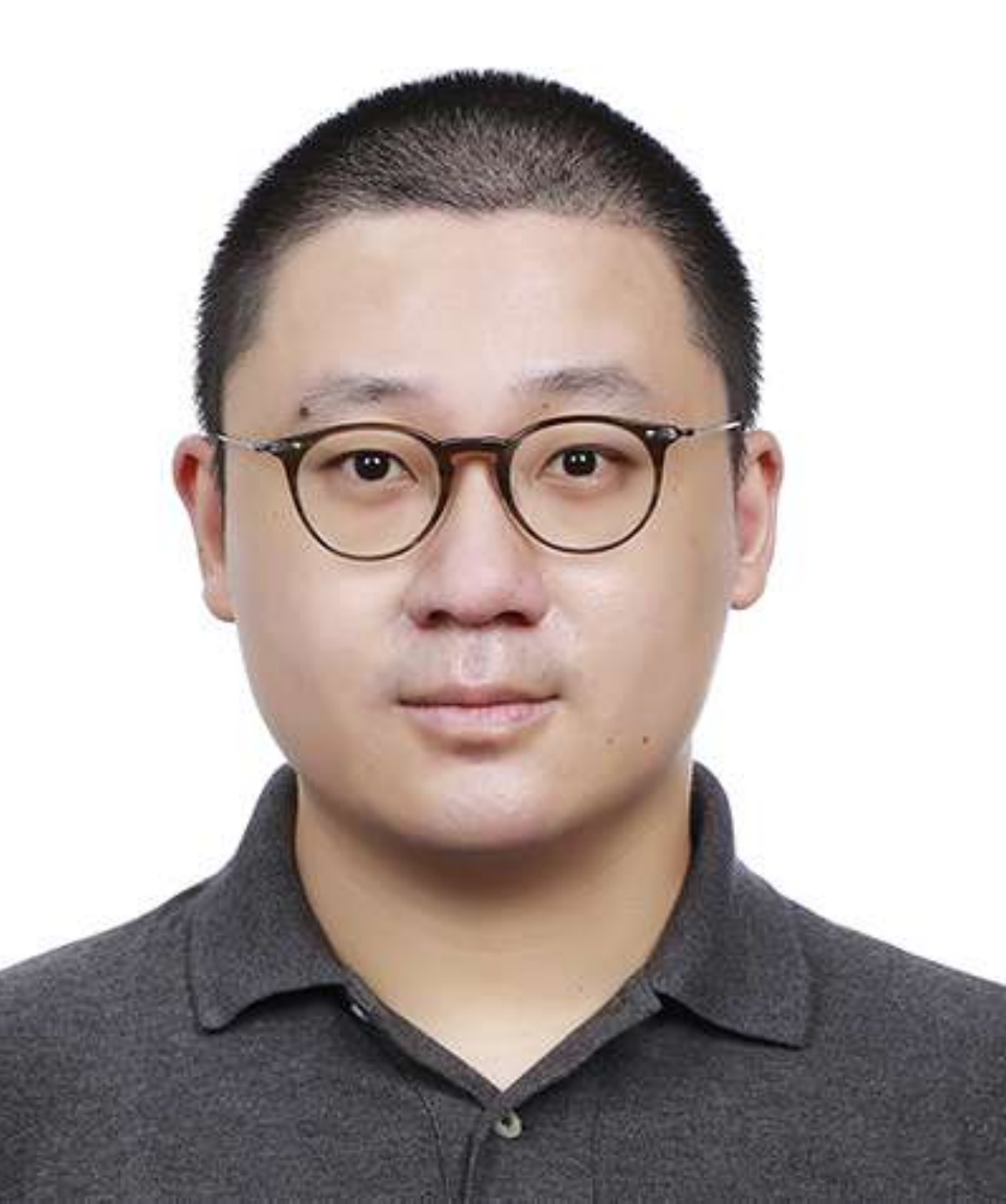}}]
{Jianyuan Bo} is currently a Ph.D. candidate at the School of Computing and Information Systems, Singapore Management University (SMU). He received his first M.S. in Mechanical Engineering from the University of Southern California, Los Angeles, United States, in 2016, followed by a second M.S. in Information Technology in Business from SMU in 2020. He earned his B.S. degree in Mechatronic Engineering from Huazhong Agricultural University, Wuhan, China, in 2014. His research interests are focused on graph representation learning and graph neural networks.
\end{IEEEbiography}

\begin{IEEEbiography}
[{\includegraphics[width=1in,height=1.25in,clip,keepaspectratio]{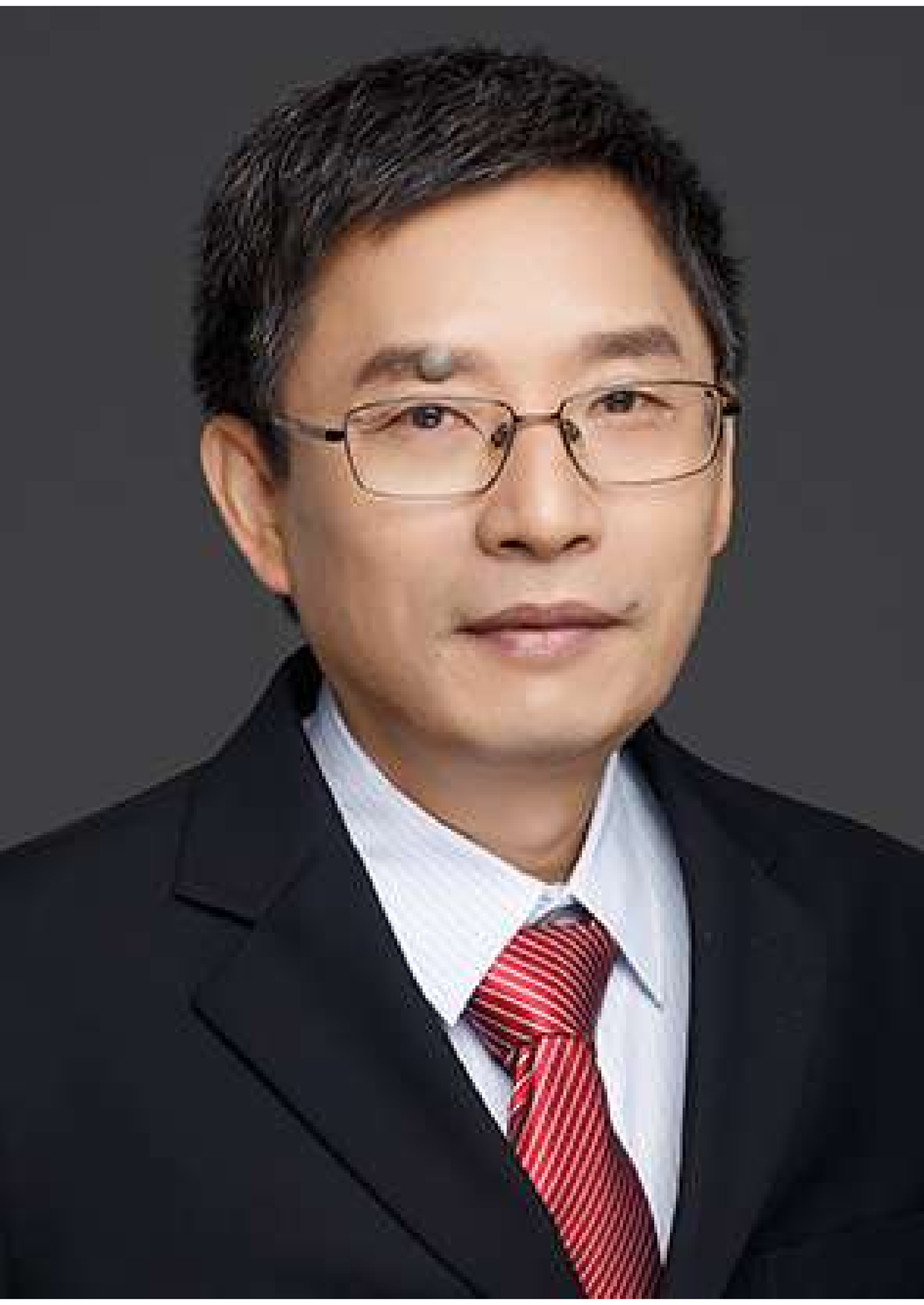}}]
{Xinming Zhang}
(Senior Member, IEEE) received the BE and ME degrees in electrical engineering from the China University of Mining and Technology, Xuzhou, China in 1985 and 1988, respectively, and the PhD degree in computer science and technology from the University of Science and Technology of China, Hefei, China in 2001. Since 2002, he has been with the faculty of the University of Science and Technology of China, where he is currently a professor with the School of Computer Science and Technology. 
\end{IEEEbiography}

\begin{IEEEbiography}
[{\includegraphics[width=1in,height=1.25in,clip,keepaspectratio]{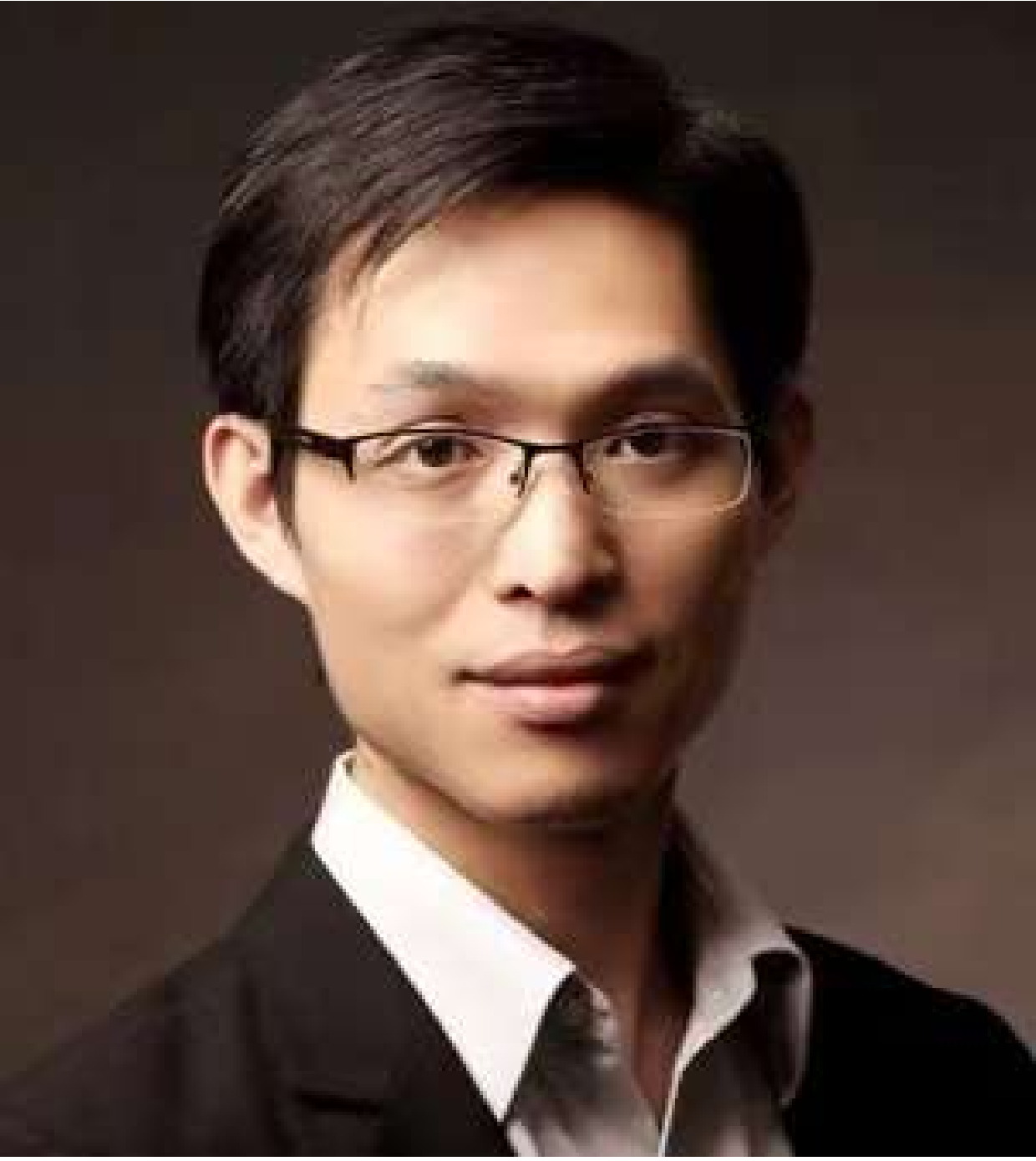}}]
{Steven C.H. Hoi}
(Fellow, IEEE) received the bachelor’s degree in computer science and technology from Tsinghua University in 2002, the master degree from Chinese University of Hong Kong in 2004 and the PhD degree in computer science from Chinese University of Hong Kong in 2006. He is currently a professor with the School of Computing and Information Systems, Singapore Management University. His current research focuses on machine Learning, deep Learning, computer vision and pattern recognition, social media and web mining.
\end{IEEEbiography}

\end{document}